\documentclass{article}

\usepackage{microtype}
\usepackage{graphicx}
\usepackage{subfigure}
\usepackage{booktabs} 

\usepackage{hyperref}

\usepackage[accepted]{icml2024}

\usepackage{amsmath}
\usepackage{amssymb}
\usepackage{mathtools}
\usepackage{amsthm}

\usepackage[capitalize,noabbrev]{cleveref}

\theoremstyle{plain}

\theoremstyle{definition}

\theoremstyle{remark}

\usepackage[textsize=tiny]{todonotes}

\icmltitlerunning{Confronting Reward Overoptimization for Diffusion Models: A Perspective of Inductive and Primacy Biases}

\begin{document}

\twocolumn[
\icmltitle{Confronting Reward Overoptimization for Diffusion Models: \\
           A Perspective of Inductive and Primacy Biases}



\icmlsetsymbol{equal}{*}

\begin{icmlauthorlist}
\icmlauthor{Ziyi Zhang}{aff1,aff2}
\icmlauthor{Sen Zhang}{aff3}
\icmlauthor{Yibing Zhan}{aff4}
\icmlauthor{Yong Luo}{aff1,aff2}
\icmlauthor{Yonggang Wen}{aff5}
\icmlauthor{Dacheng Tao}{aff5}
\end{icmlauthorlist}

\icmlaffiliation{aff1}{Institute of Artificial Intelligence, School of Computer Science, Wuhan University, China}
\icmlaffiliation{aff2}{Hubei Luojia Laboratory, Wuhan, China}
\icmlaffiliation{aff3}{The University of Sydney, Australia}
\icmlaffiliation{aff4}{JD Explore Academy, Beijing, China}
\icmlaffiliation{aff5}{Nanyang Technological University, Singapore}

\icmlcorrespondingauthor{Yong Luo}{luoyong@whu.edu.cn}

\icmlkeywords{diffusion models, text-to-image models, alignment, reinforcement learning, reward overoptimization}

\vskip 0.3in
]



\printAffiliationsAndNotice{}  

\begin{abstract}
Bridging the gap between diffusion models and human preferences is crucial for their integration into practical generative workflows. While optimizing downstream reward models has emerged as a promising alignment strategy, concerns arise regarding the risk of excessive optimization with learned reward models, which potentially compromises ground-truth performance. In this work, we confront the reward overoptimization problem in diffusion model alignment through the lenses of both inductive and primacy biases. We first identify a mismatch between current methods and the temporal inductive bias inherent in the multi-step denoising process of diffusion models, as a potential source of reward overoptimization. Then, we surprisingly discover that dormant neurons in our critic model act as a regularization against reward overoptimization while active neurons reflect primacy bias. Motivated by these observations, we propose Temporal Diffusion Policy Optimization with critic active neuron Reset (TDPO-R), a policy gradient algorithm that exploits the temporal inductive bias of diffusion models and mitigates the primacy bias stemming from active neurons. Empirical results demonstrate the superior efficacy of our methods in mitigating reward overoptimization. Code is avaliable at \url{https://github.com/ZiyiZhang27/tdpo}.
\end{abstract}

\section{Introduction}
\label{section:1}

Diffusion models \cite{d2015} represent the state-of-the-art in generative modeling for continuous data, particularly excelling in text-to-image generation \cite{sd}. Traditional training methodologies for diffusion models predominantly adhere to a maximum likelihood objective. However, such approaches may not inherently prioritize the optimization of downstream objectives, such as image aesthetic quality \cite{aes} or human preferences \cite{imagereward,hpsv2}. To align pre-trained diffusion models with downstream objectives, researchers have explored using learned or handcrafted reward functions to finetune these models. Typical solutions along this research direction can be categorized into supervised learning \cite{lee2023,hps,dong2023}, reinforcement learning (RL) \cite{dpok,ddpo}, and backpropagation through sampling \cite{draft,alignprop}.

\begin{figure}[t]
\begin{center}
\centerline{
\includegraphics[width=\columnwidth]{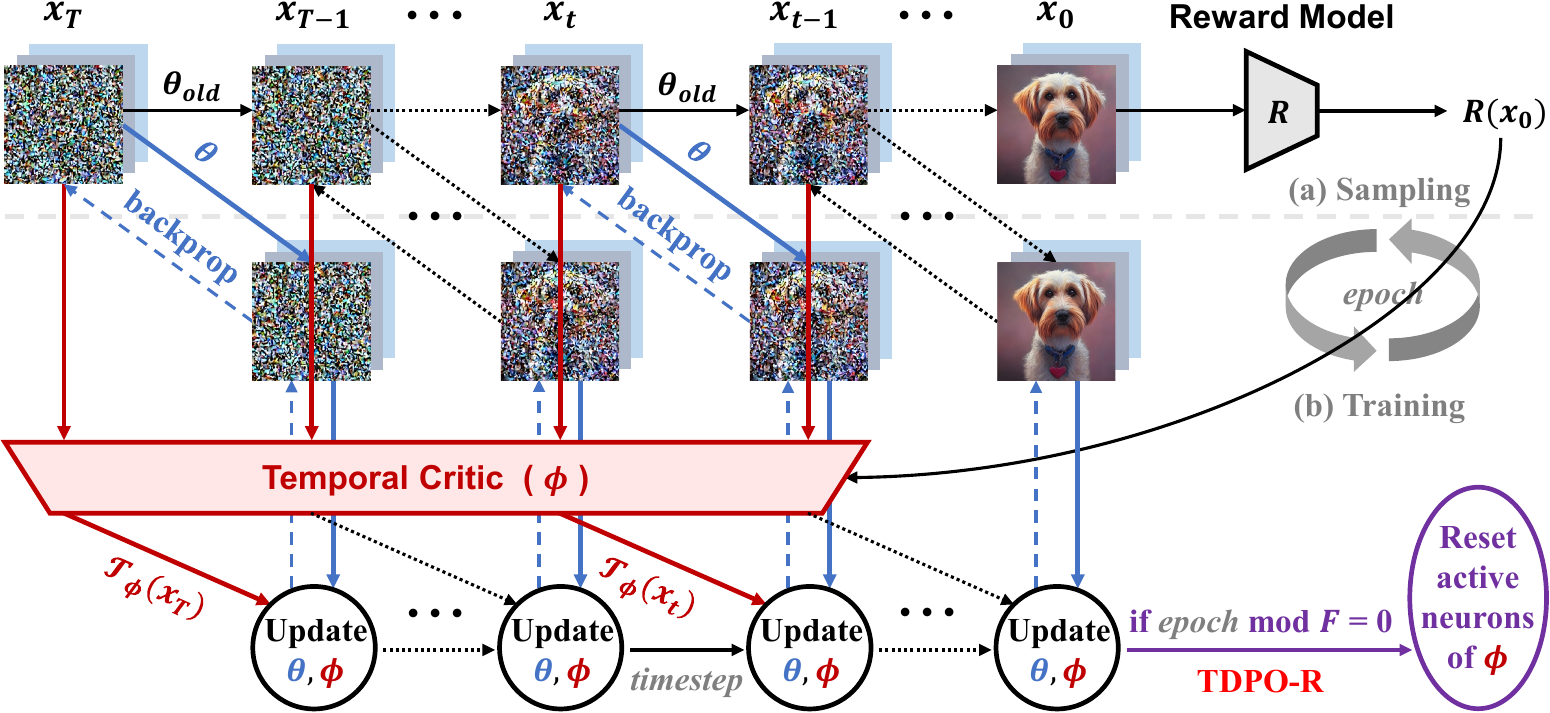}
}
\caption{TDPO-R first samples trajectories $(x_T, x_{T-1}, ... , x_0)$ from the denoising process of a fixed diffusion model parameterized by $\theta_\mathrm{old}$ for each epoch. At each timestep $t$, it performs a one-step denoising using the current diffusion model parameterized by $\theta$, estimates a temporal reward $\mathcal{T}_{\phi}(x_t)$ using a temporal critic parameterized by $\phi$, and updates the gradients for both $\theta$ and $\phi$ according to the corresponding objective functions. Additionally, TDPO-R resets active neurons of $\phi$ at the end of every $F$ epochs.}
\label{TDPO}
\end{center}
\vskip -0.2in
\end{figure}

Despite the promise of reward-driven approaches, reward overoptimization remains a fundamental and under-researched challenge. This phenomenon, characterized by overfitting learned or handcrafted reward models, stems from the inherent limitations of these models in capturing the full spectrum of human intent. In image generation, reward overoptimization typically manifests as fidelity deterioration or continual degradation in cross-reward generalization against out-of-domain reward functions. Additionally, sample efficiency further complicates this issue. Notably, while RL-based methods \cite{dpok,ddpo} exhibit relatively lower susceptibility to reward overoptimization, this advantage comes at the expense of diminished sample efficiency due to the extra sampling process isolated from training. This further entails a trade-off between sample efficiency and reward overoptimization.

Regrettably, the underlying causes of reward overoptimization in diffusion model alignment remain unclear, which is the primary concern of this work. To this end, we systematically investigate this problem from the perspective of both inductive and primacy biases. Firstly, within the context of deep RL, the consistency between the inductive bias of an algorithm and the solving task plays a crucial role in achieving robust generalization \cite{zhang2018}. However, current reward-driven alignment approaches for diffusion models exclusively focus on maximizing rewards computed from the final generated images, while overlooking the sequential nature of diffusion models and valuable intermediate information within the multi-step denoising process. This mismatch between the reward structure and the model's inherent temporal inductive bias potentially leads to overfitting and misalignment between the desired outcome (high reward) and the actual quality of the generation process.

Secondly, primacy bias \cite{primacy}, the tendency of deep RL agents to overfit early training experiences, poses another potential source of reward overoptimization. To this end, we investigate the neuron states as internal indicators of primacy bias. Although \citet{dormant} suggest that dormant neurons in deep RL agents have a negative effect on the model capacity and resetting dormant neurons reduces this effect, we surprisingly discover that dormant neurons instead act as an adaptive regularization against reward overoptimization, while active neurons appear to be susceptible to the primacy bias towards this phenomenon.

Motivated by the above observations, we propose Temporal Diffusion Policy Optimization with critic active neuron Reset (TDPO-R), a novel policy gradient algorithm that exploits the temporal inductive bias inherent in the denoising process of diffusion models and mitigates the primacy bias stemming from active neurons. As illustrated in Figure~\ref{TDPO}, to exploit the temporal inductive bias, TDPO-R assigns each intermediate denoising timestep a temporal reward, which is derived by learning a temporal critic function. The diffusion model and the temporal critic are then optimized simultaneously via gradient descent with a per-timestep update strategy. Accordingly, the consistent temporal granularity between the temporal rewards and the per-timestep gradient updates not only mitigates reward overoptimization, but also improves sample efficiency by striking a balance between update frequency and stability. To counteract the primacy bias, TDPO-R employs a periodic reset strategy that specifically targets active neurons within the critic, further alleviating the reward overoptimization problem.

We deploy the proposed TDPO-R with Stable Diffusion v1.4 \cite{sd} and conduct empirical evaluations using multiple reward functions over a variety of prompt sets. We first employ individual reward functions to quantitatively measure the sample efficiency and model performance per task. Then we introduce a novel metric of cross-reward generalization as a proxy for the quantitative evaluation of reward overoptimization. Evaluation results demonstrate the superior efficacy of our algorithms in the trade-off between sample efficiency and cross-reward generalization compared to state-of-the-art methods. In addition, we show that our critic active neuron reset strategy significantly contributes to a further mitigation of reward overoptimization during the RL process of our general training framework (i.e., TDPO), as evidenced by the outstanding performance in cross-reward generalization, as well as the fidelity and diversity observed in high-reward qualitative results. The main contributions of this paper are summarized as follows:
\begin{itemize}
\item To the best of our knowledge, this is the first work that investigates the underlying causes of reward overoptimization in diffusion model alignment from the perspective of inductive and primacy biases.
\item We exploit the temporal inductive bias of diffusion models to design TDPO, an RL-based diffusion alignment framework with consistent temporal granularity of rewards and gradients, not only mitigating reward overoptimization but also improving sample efficiency.
\item Building on TDPO, we identify the susceptibility of the critic's active neurons to primacy bias, which contributes to overoptimization, and address it with TDPO-R, which enhances TDPO with a periodic neuron reset strategy to further mitigate reward overoptimization.
\item We develop a quantitative metric of cross-reward generalization as a proxy for the evaluation of reward overoptimization, and demonstrate the superior efficacy of our methods in trading off efficiency and generalization.
\end{itemize}

\section{Related Work}

\textbf{Reward finetuning of diffusion models.} \citet{lee2023} and \citet{hps} finetune diffusion models on rewards using supervised learning. \citet{dong2023} present an online variant of these supervised learning-based methods. \citet{dpok} and \citet{ddpo} explore using policy gradient-based RL algorithms to align diffusion models with arbitrary rewards. \citet{draft} and \citet{alignprop} finetune diffusion models by backpropagating gradients of differentiable reward functions and truncate backpropagation to a few sampling steps. All these works use timestep-independent rewards based on fully-generated images, precluding intermediate samples in the denoising process. In contrast, we introduce timestep-dependent rewards for intermediate samples, and optimize diffusion models on these rewards at temporal granularity. More importantly, none of these works explicitly address reward overoptimization, which is the main focus of our work.

\textbf{Reward hacking and overoptimization.} Reward overoptimization \cite{gao2023,crlhf}, also termed ``reward hacking" \cite{skalse2022defining,inform}, refers to the detrimental phenomenon where optimizing too much on imperfect reward functions hinders the model performance on the true objectives. To address this issue, two widely employed strategies are early stopping \cite{ddpo} and Kullback-Leibler (KL) regularization \cite{dpok}. However, there still exists a lack of statistical evidence and understanding of their efficacy in reducing overoptimization. In this work, we investigate the underlying causes of reward overoptimization from the perspective of inductive and primacy biases. In addition, we are the first to design large-scale quantitative evaluations based on the cross-reward generalization metric for reward overoptimization in diffusion model alignment.

\textbf{Primacy bias and plasticity loss.} Primacy bias \cite{primacy} is also identified by a variety of other terminologies, including implicit under-parameterization \cite{kumar2020}, capacity loss \cite{lyle2021}, and dormant neuron phenomenon \cite{dormant}. All of these can be generalized as plasticity loss \cite{lyle2023,kumar2023,ma2024}, i.e., loss of ability to learn and generalize. Resetting the last layer of an agent network \cite{primacy} retrieves plasticity, but may cause knowledge forgetting. \citet{dormant} suggest that dormant neurons in agent networks have a negative effect on model plasticity, which can be mitigated by resetting these neurons. However, our empirical findings present a surprising twist in the context of reward overoptimization, which reveals that dormant neurons act as an adaptive regularization that benefits our model, offering a novel perspective on understanding neuron states and overcoming primacy bias.

\section{Preliminaries}

\subsection{Denoising Diffusion Probabilistic Models} 
This work is built upon Denoising Diffusion Probabilistic Model (DDPM) \cite{ho2020}, a well-established diffusion backbone that learns to model a probability distribution $p(x_0)$ by reversing a Markovian forward process $q(x_t | x_{t-1})$ that iteratively adds Gaussian noise towards the desired sample $x_0$ at each diffusion timestep $t$. The reverse process is modeled by a denoising neural network $\mu_{\theta}(x_t, t)$ to predict its posterior mean $\Tilde{\mu}(x_0, x_t)$, which is a weighted average of $x_0$ and $x_t$. The network is parameterized by $\theta$ and trained using the following objective:
\begin{equation}
    \resizebox{.91\linewidth}{!}{$
        \mathbb{E}_{x_0 \sim p(x_0), t \sim [1,T], x_t \sim q(x_t|x_0)} \left[\| \Tilde{\mu}(x_0, x_t) - \mu_{\theta}(x_t, t) \|^2\right].
    $}
\end{equation}%

Additionally, DDPM is readily extended to the conditional generative modeling of $p(x_0 | c)$, where $c$ is a conditional signal, such as a text prompt, processed by a conditional denoising network $\mu_{\theta}(x_t, t, c)$. To sample from a learned denoising process $p_\theta(x_{t-1} | x_t, c)$, one begins by drawing a Gaussian noisy sample $x_T \sim \mathcal{N}(0,I)$, and then employs a specific sampling scheduler \cite{ho2020,song2021}, which iteratively generates subsequent samples $x_{T-1}, ..., x_0$ according to outputs from the denoising network $\mu_{\theta}$. An intermediate timestep of the denoising process with the noise variance $\sigma^2_t$ can be written as:
\begin{equation}
    p_\theta(x_{t-1} | x_t, c) = \mathcal{N}(x_{t-1}; \mu_\theta(x_t, t, c), \sigma^2_tI).
\end{equation}%

\subsection{Reinforcement Learning} 

\textbf{Markov Decision Process (MDP).} MDP provides a mathematical framework for modeling decision-making problems. We consider finite MDP in this work, where the agent acts iteratively at each of a sequence of discrete timesteps $t \in (0, 1, 2, ...)$, up to a maximum timestep $T$. At each timestep $t$, the agent perceives a state $s_t \in S$, and selects an action $a_t \in A$ by a policy $\pi(a_t | s_t)$, where $S, A$ are state and action spaces respectively. One timestep later, the agent receives a numerical reward $r(s_t, a_t)$ as a consequence of its action, and finds itself in a new state $s_{t+1} \sim P(s_{t+1} | s_t, a_t)$, where $P$ denotes the transition probability function. 

\textbf{RL objective.} Within the MDP framework, the interaction between the agent and the environment give rise to trajectories $\tau = (s_0, a_0, s_1, a_1, ..., s_T, a_T)$, where each element represents a state-action pair at a specific timestep. Then the RL objective under this formulation is to find the policy that maximizes the expected accumulation of trajectory rewards:
\begin{equation}
    \max\limits_\pi \mathbb{E}_{\tau \sim p(\tau | \pi)}\left[ \sum_{t=0}^Tr(s_t, a_t) \right].
\end{equation}%

\section{Method}
Now we delve into our approaches to addressing the reward overoptimization problem in diffusion model alignment, focusing on the exploration of both inductive and primacy biases. First, we will address the general concern of inductive bias mismatch for reward-driven diffusion model alignment methods, by introducing a novel RL-based training framework, i.e., TDPO. Subsequently, we will investigate the primacy bias, a more specific issue within TDPO that may also contribute to reward overoptimization, and further tackle this issue by incorporating a novel periodic reset strategy for active neurons within our critic model, leading to an enhanced version of TDPO, i.e., TDPO-R.

\subsection{Temporal Diffusion Policy Optimization}
\label{section:4.1}

In this section, we aim to address the mismatch between current reward-driven alignment approaches for diffusion models and the temporal inductive bias inherent in the multi-step denoising process of diffusion models. We first extend the standard multi-step MDP formulation of the denoising process as in \cite{dpok,ddpo} by introducing timestep-dependent rewards for each denoising operation, along with an efficient approach to approximate these temporal rewards during diffusion model alignment. Building upon this new MDP formulation, we develop a novel RL framework for diffusion model alignment, i.e., Temporal Diffusion Policy Optimization (TDPO), which exploits the temporal inductive bias of the multi-step denoising process to perform temporal reward-driven optimization of diffusion polices via a per-timestep gradient update strategy.

\textbf{Temporal inductive bias.} To perform RL-based diffusion model alignment, \citet{dpok} and \citet{ddpo} map the denoising process of diffusion models to a multi-step MDP, in which the trajectories $(x_T, x_{T-1}, ... , x_0)$ correspond to the intermediate images sampled during the denoising process. In their settings, the cumulative rewards for all trajectories are condensed into a singular value $R(x_0, c)$, which is exclusively computed on the final sample $x_0$, precluding the noisy samples $x_t$ obtained at each intermediate timestep $t$. This timestep-independent reward definition creates a mismatch with the temporal inductive bias inherent in the multi-step denoising process of diffusion models, thereby posing a potential risk of overfitting to $R(x_0, c)$.

\textbf{Denoising as MDP with temporal rewards.} To inherit and exploit the temporal inductive bias within the denoising process, we characterize this process as a multi-step MDP with timestep-dependent trajectory rewards $\mathcal{T}(x_t, c)$:
\begin{equation}
    \resizebox{.91\linewidth}{!}{$
    \begin{aligned}
        & s_t \triangleq (x_{T-t}, t, c), & \rho_0(s_0) \triangleq (p(c), \delta_0, \mathcal{N}(0,I)), & \\
        & a_t \triangleq x_{T-t-1}, & P(s_{t+1} | s_t, a_t) \triangleq (\delta_c, \delta_{t+1}, \delta_{x_{T-t-1}}), & \\
        & r(s_t, a_t) \triangleq \mathcal{T}(x_{T-t-1}, c), & \pi(a_t | s_t) \triangleq p_\theta(x_{T-t-1} | x_{T-t}, c), &
    \end{aligned}
    $}
    \label{eq:mdp}
\end{equation}%
where $\rho_0$ is the initial state distribution, $\delta_z$ is the Dirac delta distribution at $z$, and optimizing the policy $\pi$ is equivalent to finetuning the diffusion model parameterized by $\theta$. This formulation diverges from the ones presented in \cite{dpok,ddpo}, in terms of the timestep-dependent definition of trajectory rewards. We refer $\mathcal{T}(x_t, c)$ to the intermediate reward w.r.t. the noisy image $x_t$ from each timestep $t$ of the denoising process.

This new MDP formulation leads to a temporal reward-driven optimization of the diffusion policy, and thus exploits the aforementioned temporal inductive bias. This optimization procedure is driven by the objective of maximizing the expected temporal rewards at each denoising timestep, i.e.,
\begin{equation}
    \max\limits_\theta \mathbb{E}_{p(c)}\mathbb{E}_{p_\theta(x_{0:T} | c)}\left[ \mathcal{T}(x_t, c) \right].
    \label{eq:obj2}
\end{equation}%

\textbf{Temporal reward approximation.} 
Prevalent reward models such as aesthetic predictor \cite{aes} and ImageReward \cite{imagereward} are usually trained on the distributions of the final clean images rather than intermediate noisy samples within the denoising process. Consequently, it is not feasible to derive the temporal rewards directly from these reward models. An intuitive solution for this problem involves retraining reward models on noisy images, but it restricts the direct utilization of off-the-shelf reward models and imposes excessive additional training overhead. 

To address this issue, we present an efficient approach in this section. In particular, we first utilize off-the-shelf reward models to compute the reward function for each final clean image $x_0$, denoted as $R(x_0, c)$. Then we approximate the temporal reward $\mathcal{T}(x_t, c)$ for each intermediate noisy image $x_t$ by learning a temporal critic function $\mathcal{T}_{\phi}(x_t, c)$ parameterized by $\phi$. To facilitate the learning process, we further use $R(x_0, c)$ as an anchor and compute $\mathcal{T}_{\phi}(x_t, c)$ as:
\begin{equation}
    \mathcal{T}(x_t, c) \approx \mathcal{T}_{\phi}(x_t, c) \triangleq R(x_0, c) - \mathcal{R}_{\phi}(x_t, c).
    \label{eq:frwd}
\end{equation}%
where $\mathcal{R}_{\phi}(x_t, c)$ is the prediction function of a temporal residual for each temporal reward, which is trained in conjunction with the policy over all denoising timesteps.

Nonetheless, learning $\mathcal{R}_{\phi}(x_t, c)$ presents a non-trivial challenge, since we need to evaluate temporal rewards at each timestep. A naive implementation of the temporal critic with a comparable number of training parameters as the reward model can incur significant training complexity. To address this challenge, we leverage the encoders of target reward models to extract embeddings from the decoded images w.r.t. each intermediate latent feature across all timesteps of the denoising process. These embeddings serve as the input to a lightweight Multi-Layer Perceptron (MLP) with only 5 linear layers, of which the final output forms the residual prediction for each temporal reward, i.e.,
%
\begin{equation}
    \mathcal{R}_{\phi}(x_t, c) = \mathrm{MLP}_{\phi}\Big(\mathrm{Encoder}_R(x_t, c)\Big).
    \label{eq:res}
\end{equation}%
\textbf{Encoder alignment.} This practice of reusing encoders establishes an alignment between the encoders of both the reward model and the temporal critic. This alignment tends to be critical for overall performance, as it ensures consistency in feature representations and enables the temporal critic to inherit the inductive bias of the reward model during initial training. This inductive bias is further dynamically refined during subsequent training, adapting to the evolving intermediate states encountered during the denoising process. Beyond performance gains, encoder alignment also offers a compelling advantage in memory efficiency, as the need to store a separate encoder for the temporal critic is eliminated. Implementation details and extended analysis of encoder alignment are provided in Appendix~\ref{appendix:encoder}.

\textbf{Temporal gradient estimation.} Given the temporal rewards in Eq.~(\ref{eq:frwd}) and (\ref{eq:res}), we can estimate the gradient of the objective in Eq.~(\ref{eq:obj2}) w.r.t. the policy parameters $\theta$ at temporal granularity. We first sample trajectories $(x_T, x_{T-1}, ... , x_0)$ from the denoising process $p_\theta(x_{0:T} | c)$, and collect the likelihood gradients with respect to $\theta$, i.e., $\nabla_\theta{p_\theta(x_{t-1} | x_t, c)}$. To reuse the trajectories sampled by an old policy parameterized by $\theta_\mathrm{old}$, we employ importance sampling \cite{kakade2002}, which reweights the temporal rewards by the corresponding probability ratio. Then the temporal gradient at each denoising timestep reads:
\begin{equation}
    \mathbb{E}_{p(c)}\mathbb{E}_{p_\theta(x_{0:t} | c)}\left[ -\mathcal{T}_{\phi}(x_t, c)\nabla_\theta\frac{p_{\theta} (x_{t-1} | x_t, c)}{p_{\theta_\mathrm{old}}(x_{t-1} | x_t, c)} \right].
\end{equation}%
The temporal critic is optimized by the objective below:
\begin{equation}
    \mathbb{E}_{p(c)}\mathbb{E}_{p_\theta(x_{0:t} | c)}\left[ \left( \hat{\mathcal{R}}_{\phi}(x_t, c) - R(x_0, c) \right)^2 \right],
\end{equation}%
where $\hat{\mathcal{R}}$ denotes the new residual prediction computed during the training phase and is used to estimate gradients for the critic model, while the old residual prediction $\mathcal{R}$ in Eq.~(\ref{eq:frwd}) is computed during the sampling phase with no gradient and is used to estimate temporal rewards.

\textbf{Per-timestep gradient update.} We concurrently update the policy parameters $\theta$ and the critic parameters $\phi$ via gradient descent. In particular, we perform each update in a per-timestep manner, in contrast to other methods that employ per-batch updates, emphasizing the temporal granularity of our approach. A per-timestep gradient update for $\theta$ or $\phi$ within our general RL-based training framework (i.e., TDPO) is performed via the averaged batch gradient below:
\begin{equation}
    \frac{1}{B}\sum_{i=1}^{B}\nabla_\alpha\mathrm{G}_\alpha(x^i_t, c_i), \alpha \in \{\theta, \phi\}.
    \label{eq:objtime}
\end{equation}%
where $B$ is the batch size, and $\nabla_\alpha\mathrm{G}_\alpha(x^i_t, c_i)$ is the objective gradient estimate with respect to $\theta$ or $\phi$ at each timestep $t$ in each mini-batch $i$. The motivation and advantages of this per-timestep update strategy are described as follows:

\textbf{Remark on per-timestep update.} In most cases of deep RL, a higher gradient update frequency often results in faster convergence but worse stability. In our settings, we operate on samples spanning two dimensions: timesteps and mini-batches, allowing us to elevate the gradient update frequency by reducing the sizes of these dimensions. However, reducing the dimension sizes introduces lower variance in sample distributions, potentially leading to overfitting. Intuitively, the variance of sample distributions within a per-timestep context (encompassing all mini-batches) exceeds that within a per-minibatch context (covering all timesteps derived from a shared Gaussian distribution). This suggests that reducing the number of timesteps per update represents a relatively secure approach for expediting convergence, while it still poses a risk of overfitting to the timestep-independent reward exclusively based on the final image. To mitigate this risk, our TDPO incorporates the temporal rewards as fine-grained guidance for the per-timestep updates, thereby improving sample efficiency while ensuring overall stability.

\begin{figure*}[t]
\vskip 0.2in
\begin{center}
\centerline{
\includegraphics[width=0.88\textwidth]{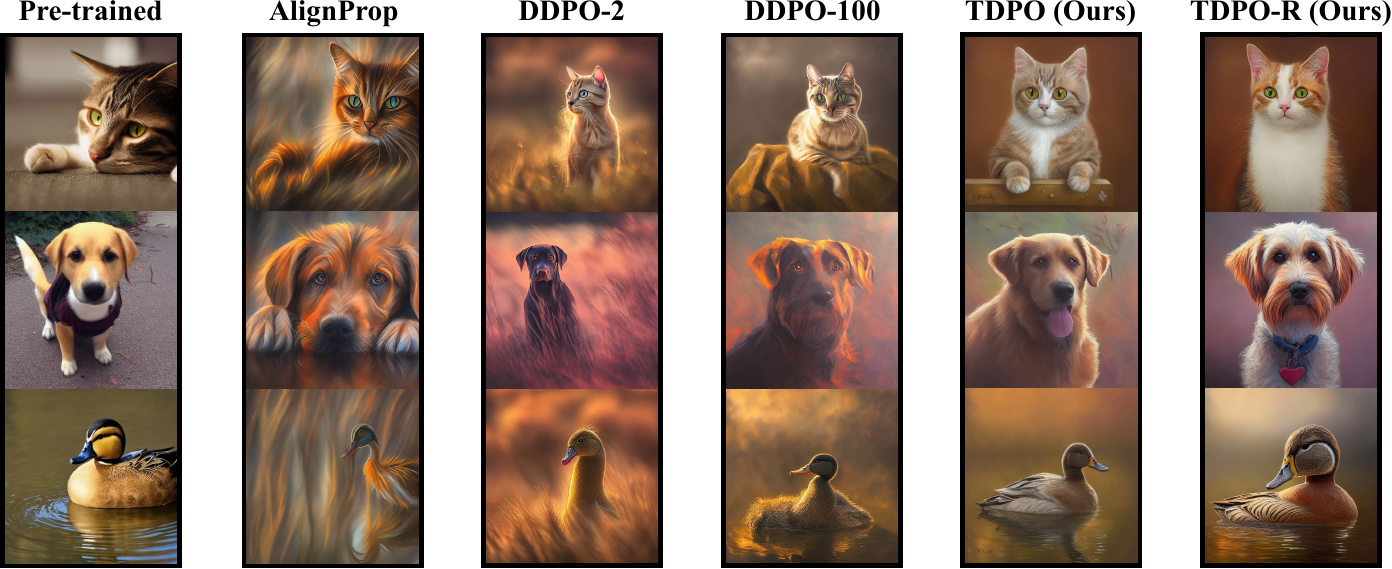}
}
\caption{Image generation results sampled from models that are either pre-trained or further finetuned on Aesthetic Score via AlignProp, DDPO-2, DDPO-100, as well as our TDPO and TDPO-R. For a fair comparison, all images are generated using a fixed random seed of 42. Additionally, for the fine-tuned models, the aesthetic scores of the generated images achieve similar values around 7 $\pm$ 0.1.}
\label{vis}
\end{center}
\vskip -0.2in
\end{figure*}

\subsection{Primacy Bias within TDPO}
\label{section:4.2}

While TDPO mitigates reward overoptimization by incorporating the temporal inductive bias of diffusion models, primacy bias, a more specific factor contributing to reward overoptimization, may arise due to the limited model capacity of the temporal critic in the TDPO framework. In this section, we investigate how the state of neurons in our model reflects such primacy bias and how it contributes to reward overoptimization during diffusion model alignment.

\begin{algorithm}[tb]
    \caption{TDPO-R: Temporal diffusion policy optimization (with critic active neuron reset)}
    \label{alg:TDPO}
    \begin{algorithmic}
        \STATE {\bfseries Input:} Diffusion model parameters $\theta$, critic model parameters $\phi$, context distribution $p(c)$, epochs $E$, denoising timesteps $T$, batch size $B$, and neuron reset frequency $F$
        \FOR{$e = 1, ..., E$}
        \STATE Obtain samples $\{c_i \sim p(c)$, $x_{0:T}^i \sim p_\theta(x_{0:T}|c_i)\}_{i=1}^{B}$
        \STATE Compute temporal rewards according to Eq.~(\ref{eq:frwd})
            \FOR{$t = T, ..., 1$}
                \STATE Update $\theta, \phi$ at timestep $t$ according to Eq.~(\ref{eq:objtime})
            \ENDFOR
            \IF{$e \mod F == 0$}
                \STATE Obtain neuron masks for $\phi$ according to Eq.~(\ref{eq:mask})
                \STATE Reinitialize $\phi$ where the neuron mask is $true$
            \ENDIF
        \ENDFOR
        \STATE {\bfseries Output:} Optimized diffusion model parameters $\theta$
    \end{algorithmic}
    \label{algorithm:1}
\end{algorithm}

\textbf{Neuron activations and states.} Following \citet{dormant}, we use neuron activation in deep neural networks to categorize the states of neurons. We first consider each feature map between two layers within a network module as the state of a neuron. For an input $x$ of distribution $\mathcal{D}$, we denote the activation of a neuron $n$ in a module $m$ as $a_n^m(x)$, and compute an activation score $\mathcal{A}_n^m$ for each neuron $n$ in the module $m$ as: 
\begin{equation}
    \mathcal{A}_n^m = \frac{\mathbb{E}_{x \in \mathcal{D}}|a_n^m(x)|}{\frac{1}{N^m}\sum_{n \in N^m} \mathbb{E}_{x \in \mathcal{D}}|a_n^m(x)|},
    \label{eq:act}
\end{equation}%
where $N^m$ is the number of neurons in the module $m$, and the expected activation over $\mathcal{D}$ is normalized by the average of all activations. Then we set a threshold for this activation score to categorize all neurons in our models into two opposite states. If the activation score of a neuron is above the threshold, we say the neuron is active, otherwise it is regarded as dormant.

\textbf{Dormant neurons are indispensable.} Accordingly, we conduct empirical evaluations to investigate the effects of different neuron states on reward overoptimizaiton during the training process of TDPO. We detect the percentage of dormant neurons in our critic model, and observe a slow ascent of this percentage during training. 
To directly influence the neuron states during training, we periodically reset neurons of a given state by reinitializing its parameters to the original distributions. Surprisingly, we find that while resetting dormant neurons periodically reduces the dormant percentage at all training steps, it actually exacerbates reward overoptimizaiton. This deviates from the conclusion in \cite{dormant} where dormant neurons in deep RL were found to hinder model capacity and necessitate resets.

\textbf{Active neurons reflect primacy bias.} We further explore resetting active neurons in the critic. Although this does not reduce the dormant percentage, it effectively mitigates reward overoptimizaiton. Interestingly, resetting all neurons in the critic also exacerbates reward overoptimization, albeit to a lesser degree compared to solely resetting dormant neurons. We posit that dormant neurons in the critic model act as an adaptive regularization mechanism against overoptimization to imperfect rewards, which suggests resetting dormant neurons may damage this implicit regularization. Our findings imply that, within the context of reward overoptimization, primacy bias manifests primarily in active neurons. Consequently, periodically resetting these neurons offers a potential mitigation strategy, encouraging the model to learn new regularization patterns without forgetting crucial past regularization. Further details and analyses supporting this observation are provided in Section~\ref{section:5.3} and Appendix~\ref{appendix:neuron}.

We further analyze the effect of the neuron states in the policy model. Since we adopt Low-Rank Adaptation (LoRA) \cite{lora} for the policy, only neurons within the LoRA layers can be reset. We detect very few dormant neurons, and resetting them makes no significant difference to the results. In this case, resetting active neurons causes catastrophic forgetting and heavily hinders learning.

\textbf{TDPO with critic active neuron Reset (TDPO-R).} 
Motivated by the above analyses, we present TDPO-R, a variant of TDPO that periodically resets the active neurons in the critic model with a frequency $F$. In practice, to reinitialize the model parameters $\phi_a$ corresponding to the active neurons, we compute a neuron mask for each module $m$:
\begin{equation}
    \mathrm{Mask}^m = \left[\mathcal{A}_n^m > 0\right]_{n=1}^{N^m},
    \label{eq:mask}
\end{equation}%
where each boolean value in it is set to true if the corresponding activation score $\mathcal{A}_n^m > 0$. This neuron mask is used to reinitialize the weights of both the incoming and outgoing layers corresponding to the active neurons in module $m$. The pseudo-code of TDPO-R is summarized in Algorithm~\ref{algorithm:1}.

\textbf{In neuroscience,} there are several studies \cite{neuron2018, neuron2019, neuron2022} investigating the function of dormant neurons. According to \citet{neuron2019}, most neurons in the brain do not fire action potentials and remain dormant for a long time. These dormant neurons are formed during evolution, but are far from the scope of natural selection as they avoid regular functional tasks. However, under the influence of stress and disease, they occasionally become active, which can lead to various neurological and psychological disease symptoms and behavioral abnormalities. Interestingly, this conclusion mirrors our observation that resetting dormant neurons could be harmful for mitigating reward overoptimization.

\begin{figure*}[t]
\vskip 0.1in
\begin{center}
\centerline{
\includegraphics[width=0.315\textwidth]{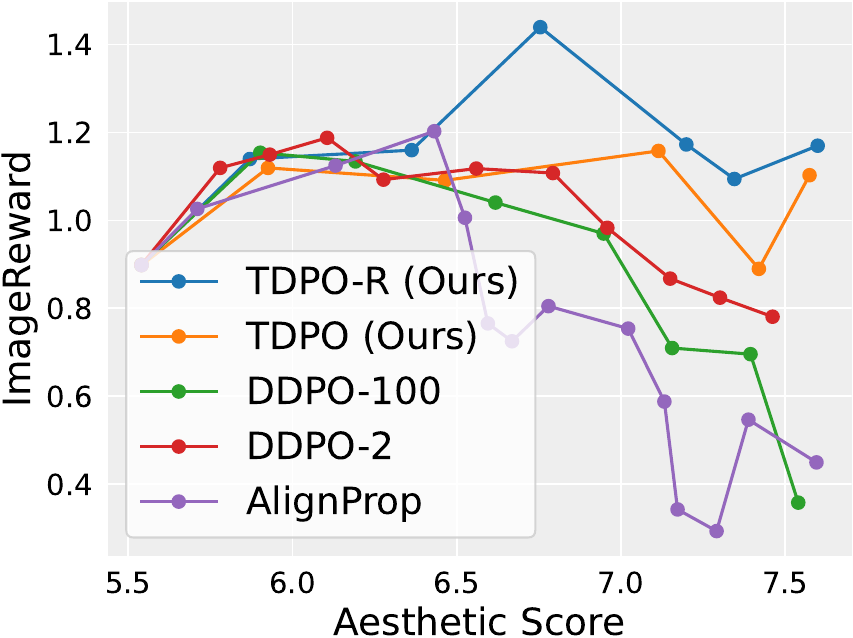}
\hskip 0.05in
\includegraphics[width=0.335\textwidth]{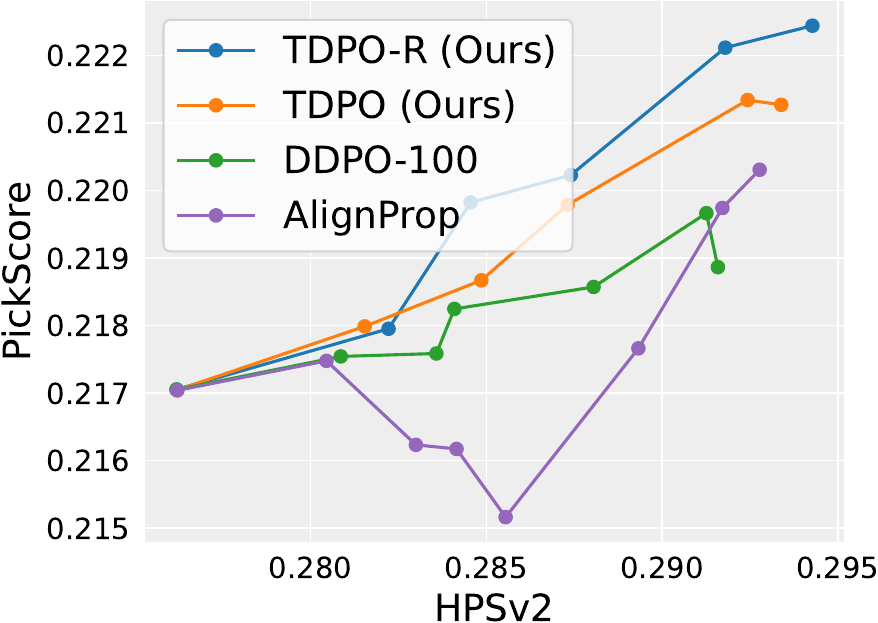}
\includegraphics[width=0.335\textwidth]{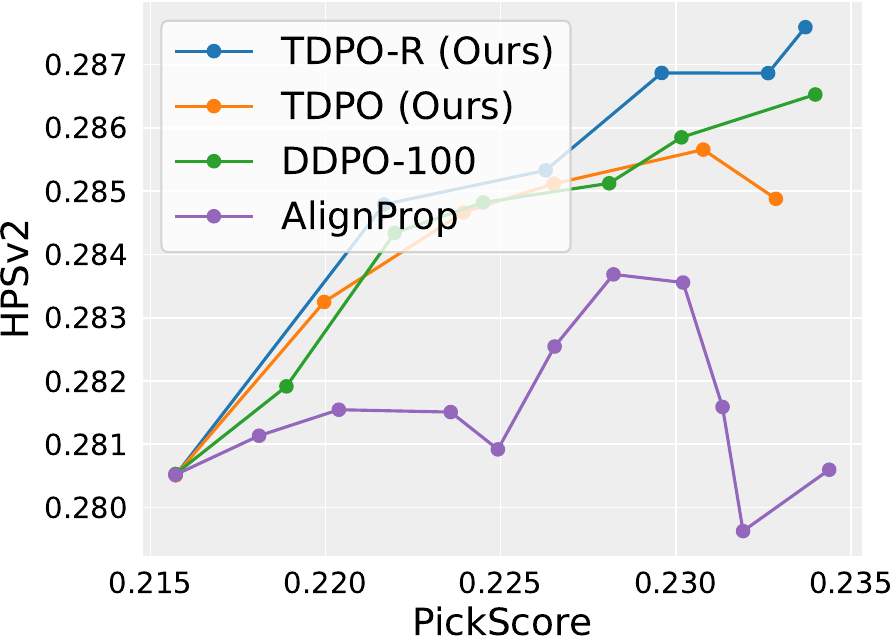}
}
\caption{Out-of-domain evaluation results via cross-reward generalization against ImageReward (left), PickScore (middle), and HPSv2 (right) when finetuning the diffusion model on Aesthetic Score (left), HPSv2 (middle), and PickScore (right), respectively.}
\label{cross}
\end{center}
\vskip -0.2in
\end{figure*}

\begin{figure*}[t]
\vskip 0.1in
\begin{center}
\centerline{
\includegraphics[width=0.34\textwidth]{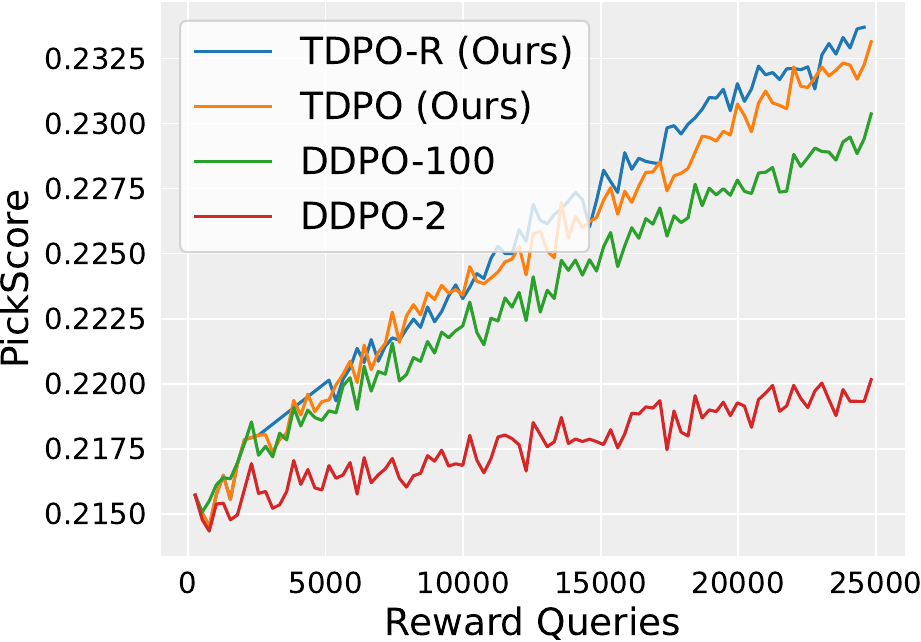}
\includegraphics[width=0.335\textwidth]{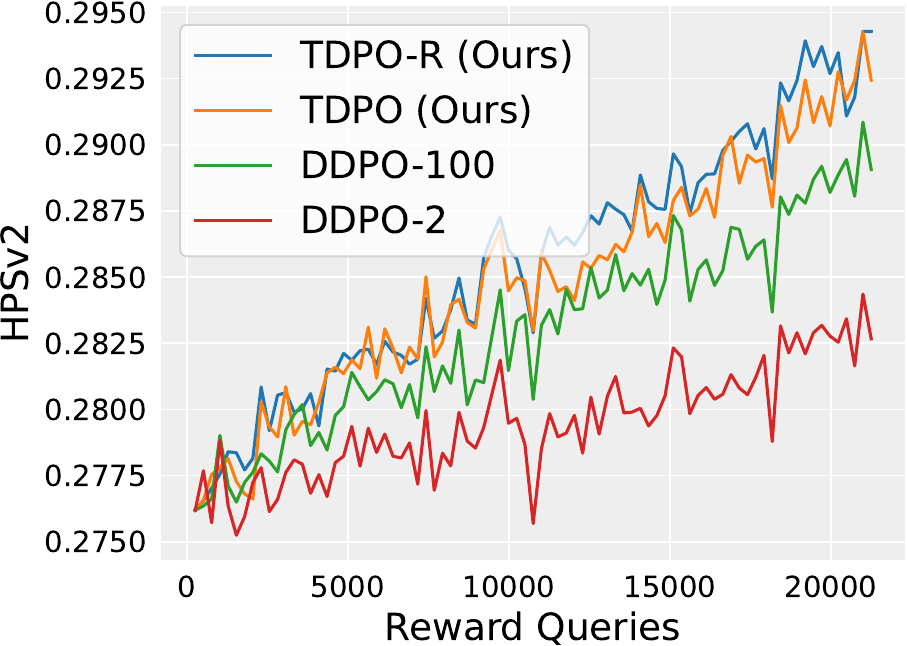}
\includegraphics[width=0.32\textwidth]{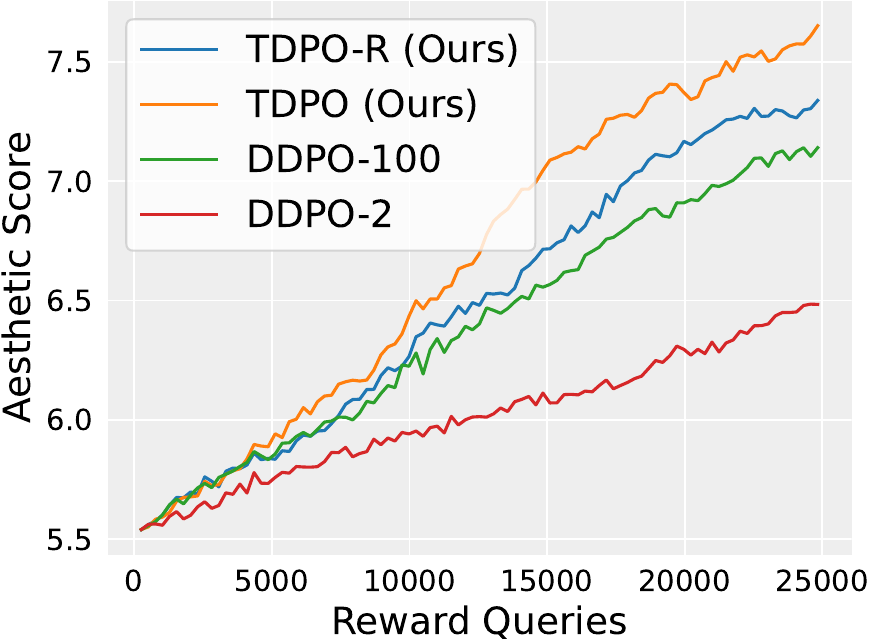}
}
\caption{Quantitative evaluation results for the efficacy of our TDPO and TDPO-R in improving sample efficiency when finetuning the diffusion model on the reward functions of PickScore (left), HPSv2 (middle), and Aesthetic Score (right), compared to DDPO with the update frequencies of 2 (DDPO-2) and 100 (DDPO-100) per epoch.}
\label{sample}
\end{center}
\vskip -0.2in
\end{figure*}

\section{Empirical Evaluations}
\label{section:5}

We conduct comprehensive experiments to validate the efficacy of our algorithms on both sample efficiency and reward overoptimization alleviation when aligning text-to-image diffusion models with diverse reward functions.

\subsection{Implementation Details}

\textbf{Baselines.} We compare our algorithms with state-of-the-art baselines\footnote{Additional baselines, including DRaFT \cite{draft} that lacks open-source code, and DPOK\cite{dpok} and ReFL\cite{imagereward} that underperform DDPO and AlignProp, are omitted in our reproductions due to the resource constraints.}, including pre-trained Stable Diffusion \cite{sd}, Denoising Diffusion Policy Optimization (DDPO) \cite{ddpo}, and AlignProp \cite{alignprop}. We use the official PyTorch codebase of DDPO for result reproduction. Following DDPO, we use Stable Diffusion v1.4 as the base generative model. For a fair comparison, we reproduce AlignProp using Stable Diffusion v1.4, while their reported results are based on v1.5.

\textbf{Reward functions.} To demonstrate the generalizability of our method, we perform training and evaluation on diverse reward functions, in which
\textbf{(a) Aesthetic Score} is computed using the LAION aesthetic predictor \cite{aes} and a text prompt set consisting of 45 animal names consistent with that in \cite{ddpo};
\textbf{(b) PickScore} \cite{pick} is employed as an objective reward function for human preference learning, using the same prompt set as for Aesthetic Score;
\textbf{(c) Human Preference Score v2} (HPSv2) \cite{hpsv2} presents another reward function for human preference learning, along with 802 prompts drawn from Human Preference Dataset v2;
\textbf{(d) ImageReward} \cite{imagereward} is employed as an evaluation score function for cross-reward generalization.

To establish a consistent benchmark, the training procedures and configurations of our algorithms are based on the official PyTorch implementation of DDPO, which adopts LoRA \cite{lora} to reduce memory and computation cost.

\subsection{Sample Efficiency in Diffusion Model Alignment}
\label{section:5.2}

We employ TDPO and TDPO-R to separately finetune Stable Diffusion v1.4 on Aesthetic Score, PickScore, and HPSv2. We report the average reward over samples at each training interval w.r.t. specific number of reward queries, as the indicator of sample efficiency. TDPO(-R) performs each gradient update in a per-timestep manner with all batch samples averaged, resulting in a higher update frequency (100 gradient updates per epoch) compared to the original DDPO implementation (2 gradient updates per epoch). Thus, for a direct comparison, we further reproduce DDPO using the same update frequency as ours. Figure~\ref{sample} shows that our algorithms consistently outperform both implementations of DDPO on each of the three rewards, demonstrating their effectiveness in improving sample efficiency. Notably, the high-frequency updates also accelerate training for DDPO, albeit at the cost of exacerbating reward overoptimization, as highlighted in Figure~\ref{cross} and further discussed in Section~\ref{section:5.3}.

\subsection{Reward Overoptimization and Generalization}
\label{section:5.3}

\textbf{Cross-reward generalization.} To quantitatively assess reward overoptimization, we introduce cross-reward generalization, where the model is evaluated against out-of-domain reward functions after being finetuned on a specific reward function. As shown in Figure~\ref{cross}, the X-axis represents the training objective reward, while the Y-axis represents the evaluation score calculated by out-of-domain reward functions. Reward overoptimization typically leads to a decline or slow rise in the evaluation score as the training reward increases. In Figure~\ref{cross}, TDPO and TDPO-R exhibit superior performance in three sets of cross-reward evaluations compared to DDPO and AlignProp, demonstrating the effectiveness of our methods in mitigating reward overoptimization.

\begin{figure}[ht]
\vskip 0.1in
\begin{center}
\centerline{
\includegraphics[width=\columnwidth]{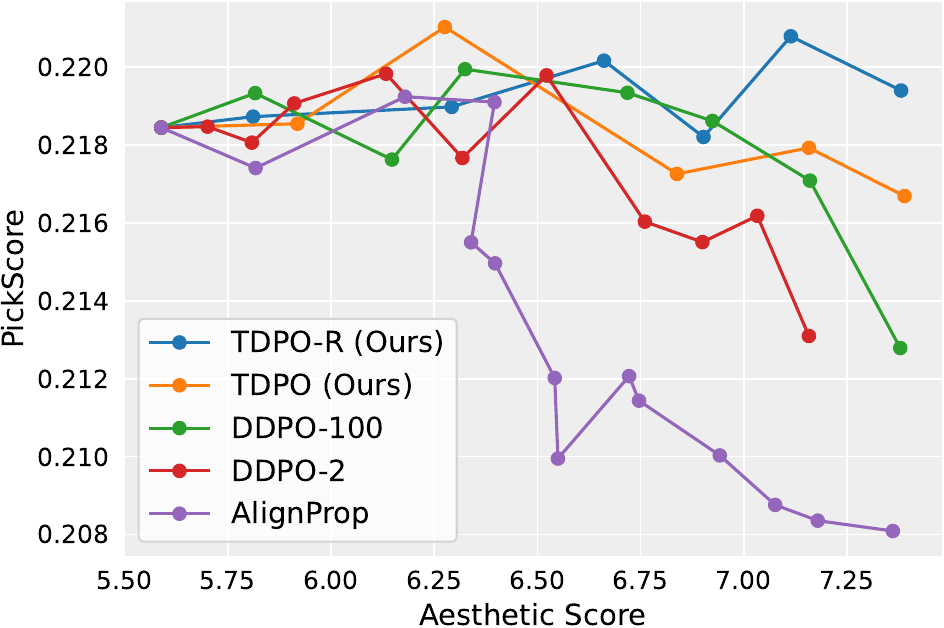}
}
\caption{Cross-reward generalization results evaluated on a text prompt set of unseen animals when finetuning the diffusion model on Aesthetic Score.}
\label{unseen}
\end{center}
\vskip -0.2in
\end{figure}

\textbf{Generalization to unseen prompts.} 
As shown in Figure~\ref{unseen}, TDPO and TDPO-R maintain their superior performance of cross-reward generalization even on novel text prompts unseen during the finetuning process, which further emphasizes their effectiveness and robustness against reward overoptimization. Implementation details and more evaluation results on unseen prompts are provided in Appendix~\ref{appendix:unseen}.

\begin{figure}[ht]
\vskip 0.1in
\begin{center}
\centerline{
\includegraphics[width=\columnwidth]{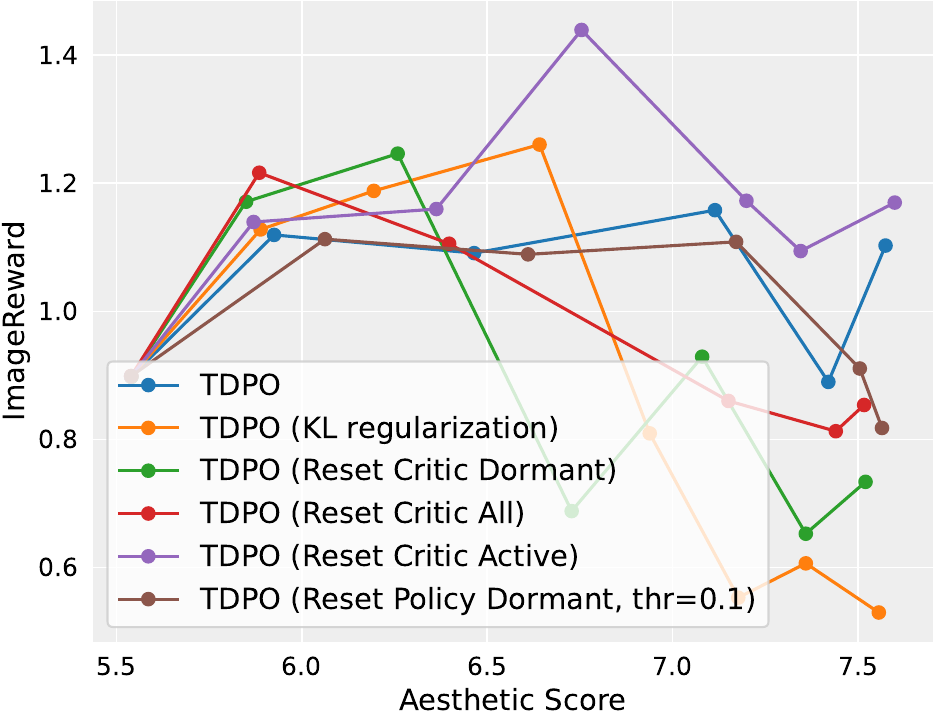}
}
\caption{Evaluation results of cross-reward generalization to ImageReward when finetuning the diffusion model on Aesthetic Score, comparing different variants of TDPO.}
\label{aes_img_reset}
\end{center}
\vskip -0.2in
\end{figure}

\textbf{Effects of neuron states.} As mentioned in Section~\ref{section:4.2}, we investigate the effects of neuron states on reward overoptimizaiton by comparing TDPO and its variants with different reset strategies. In Figure~\ref{aes_img_reset}, we present the results of cross-reward generalization to ImageReward for the variants that reset all, dormant, and active neurons in our critic model with the activation score threshold set to 0. For the variant that resets dormant neurons in our policy model, we set the threshold to 0.1, as using a threshold of 0 shows negligible differences from the standard TDPO. The results are consistent with the effects discussed in Section~\ref{section:4.2}. Further analyses and evaluation results are provided in Appendix~\ref{appendix:neuron}.

\textbf{Alternate strategies for overoptimization.} Both DDPO and AlignProp apply early stopping to prevent overoptimization, but the reliance on interactive inspection hinders its scalability. \citet{dpok} conduct an analysis of KL regularization in diffusion model alignment, while the quantitative evaluation on reward overoptimization is lacking. We also report the cross-reward generalization results of using KL regularization in Figure~\ref{aes_img_reset}, which indicates that our neuron reset strategy is more effective for mitigating reward overoptimization compared to KL regularization.

\begin{figure*}[ht]
\vskip 0.2in
\begin{center}
\centerline{
\includegraphics[width=\textwidth]{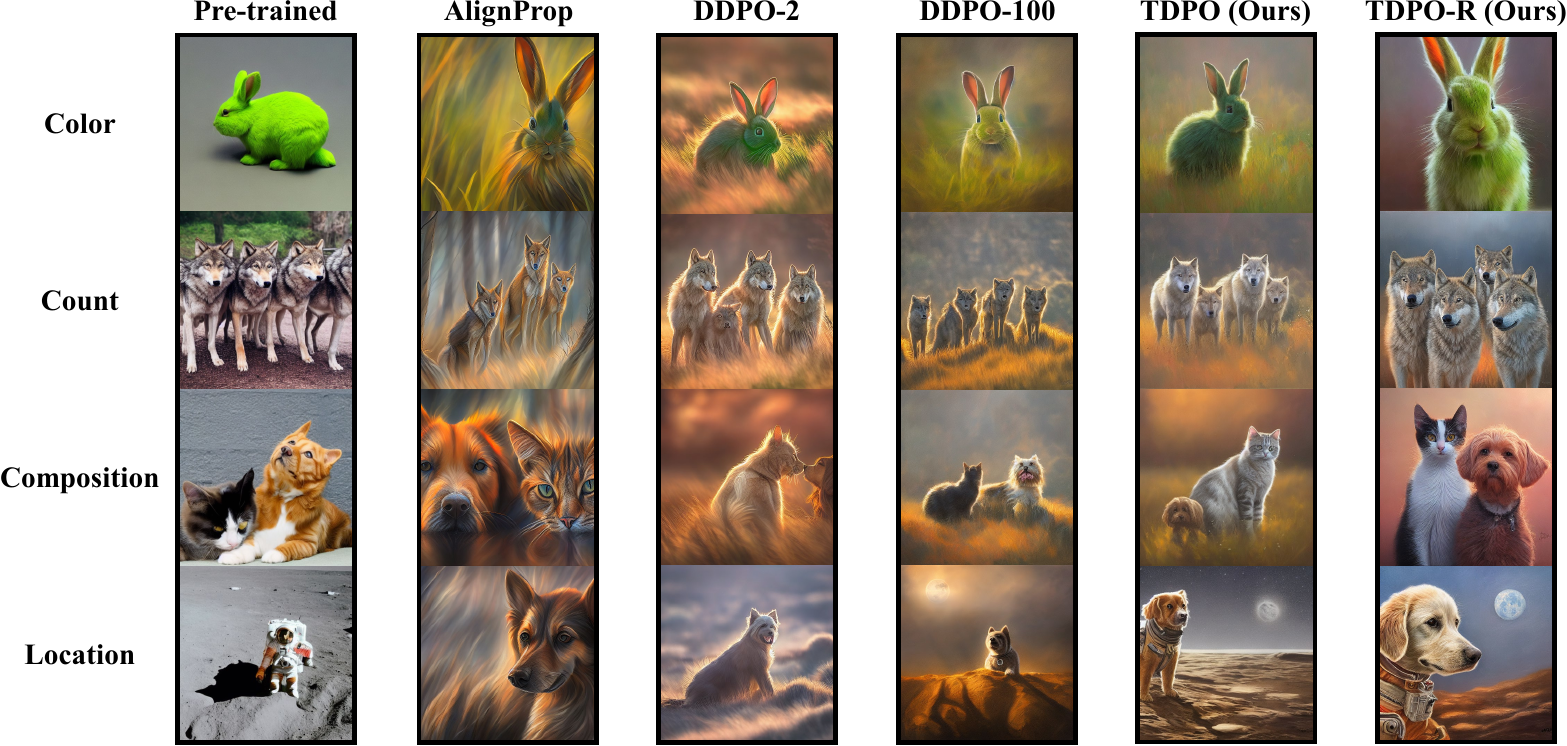}
}
\caption{Image generation results for unseen text prompts involving color (``A green colored rabbit"), count (``Four wolves in the park"), composition (``A cat and a dog"), and location (``A dog on the moon") from models either pre-trained or further finetuned on Aesthetic Score. For a fair comparison, all images are generated using a fixed random seed of 42. Additionally, for the fine-tuned models, the aesthetic scores of the generated images achieve similar values around 7 $\pm$ 0.1.}
\label{vis_complex}
\end{center}
\vskip -0.2in
\end{figure*}

\subsection{Qualitative Comparison}
\label{section:5.4}
In Figure~\ref{vis}, we compare the high-reward image results of the alignment methods when optimizing rewards to the same degree. The results from AlignProp and DDPO show notable saturation in terms of style, background, and sunlight, while our results manifest greater diversity in these aspects and exhibit higher fidelity, which highlights the effectiveness of our methods in mitigating reward overoptimization. Furthermore, we also provide additional qualitative results for unseen text prompts in Figure~\ref{vis_complex}. In comparison with other methods, our results are better aligned with the prompts in terms of color, count, composition, and location, and also exhibit higher image fidelity, indicating a lower degree of reward overoptimization.

\section{Conclusion}

In this work, we confront reward overoptimization in diffusion model alignment from the perspective of inductive and primacy biases. Specifically, we identify the temporal inductive bias of diffusion models and surprisingly discover that active neurons in our proposed temporal critic reflect the primacy bias. Inspired by these findings, we present TDPO-R, which exploits the temporal inductive bias of diffusion models and addresses the primacy bias during its RL training process. Empirical evaluations validate the effectiveness of the proposed methods in mitigating reward overoptimization.

\textbf{Limitations and future work.} To address computational limitations, our TDPO-R adopts LoRA finetuning instead of full model finetuning for diffusion models, precluding a comprehensive analysis of the diffusion model's internal neuron states in this case. However, this work opens avenues for follow-up research on reward overoptimization of diffusion models. Moreover, the potential of multi-reward learning for diffusion models remains under-explored, highlighting a significant gap for future work. We hope that our work will also inspire further exploration of potential reward overoptimization in this new domain of multi-reward learning for diffusion models.

\section*{Acknowledgements}
This work is supported in part by the Major Science and Technology Innovation 2030 ``New Generation Artificial Intelligence'' key project (No. 2021ZD0111700), the National Natural Science Foundation of China (Grant No. U23A20318 and 62276195), the Special Fund of Hubei Luojia Laboratory under Grant 220100014, and the National Research Foundation Singapore and DSO National Laboratories under the AI Singapore Programme (AISG Award No: AISG2-GC-2023-006).
Dr. Tao's research is partially supported by NTU RSR and Start Up Grants.

\section*{Impact Statement}

This work contributes to the advancement of diffusion model alignment, with the potential to impact various aspects of society. Here, we highlight the key positive impacts:

\textbf{Improved alignment of diffusion models.} This work confronts the issue of reward overoptimization, which hinders the effective alignment of diffusion models with downstream applications. By mitigating this issue, we pave the way for the development of more reliable and trustworthy diffusion models that reflect human preferences, which empowers individuals and businesses to leverage the powerful capabilities of diffusion models for various creative applications.

\textbf{Potential for broader applications.} Beyond their direct impacts on diffusion models, the insights and techniques presented in this work, such as exploiting temporal inductive bias and addressing primacy bias through active neuron reset, may hold broader applicability in other domains of deep reinforcement learning, where similar challenges of overoptimization and bias hinder effective learning.

Furthermore, it is also essential to acknowledge potential societal concerns associated with this technology, such as:

\textbf{Misuse of diffusion models.} As diffusion models evolve towards enhanced alignment with human preferences and increased controllability, concerns regarding their potential misuse for malicious purposes, such as generating discriminatory or harmful contents, become increasingly salient. It is crucial to develop safeguards and ethical guidelines alongside technological advancements to mitigate these risks.

\textbf{Unintended biases in reward learning.} The effectiveness of diffusion model alignment relies on accurately capturing human preferences from reward models. However, human preferences can be subjective and biased. It's crucial to consider and mitigate potentially unintended biases in the data used to train the reward model to avoid exploiting and amplifying such biases in the generated outputs.

\nocite{gae}
\nocite{ppo}

\bibliography{icml2024}
\bibliographystyle{icml2024}

\newpage
\appendix
\onecolumn

\section{Additional Implementation Details}
\label{appendix:imp}

In all experiments, we use Stable Diffusion v1.4 \cite{sd} as the base generative model, which ensures consistency with DDPO \cite{ddpo} and allows for a direct comparison with AlignProp \cite{alignprop}, despite their use of v1.5 in AlignProp. In addition, we conduct diffusion model alignment on the LoRA weights \cite{lora} of the U-Net architecture instead of the full parameter set to reduce memory and computation overheads, following established practices and aligning with official implementations of both DDPO and AlignProp.

\textbf{DDPO implementations.} We use the official PyTorch codebase of DDPO for result reproduction. As discussed in Section~\ref{section:5.2}, our TDPO(-R) performs each gradient update in a per-timestep manner with all batch samples averaged, resulting in a higher update frequency (100 gradient updates per epoch) compared to the original DDPO implementation (2 gradient updates per epoch). For a fair comparison, we further reproduce DDPO using the same update frequency (100) and learning rate (1e-4) as ours. Since the PyTorch implementation of DDPO adopts gradient accumulation to reach larger effective batch sizes without requiring additional memory, we adjust its update frequency by reducing the gradient accumulation steps per epoch from 16$\times T$ to 16, where $T$ denotes the number of denoising timesteps with a default value of 50. This leads to two variants of DDPO implementation: DDPO-2 and DDPO-100, differing exclusively in the hyperparameters governing the update frequency and the learning rate. All experiments were conducted on a system equipped with 8 NVIDIA A100 GPUs with 40GB of memory each.

\textbf{AlignProp implementation.} To facilitate a direct comparison, we reproduce AlignProp using Stable Diffusion v1.4, while their reported results are based on v1.5. All experiments were conducted on a system equipped with 4 NVIDIA A100 GPUs with 40GB of memory each, adhering to their default configurations, with an exception of the base model version.

\textbf{TDPO and TDPO-R implementations.} For consistency, the training procedures and configurations of our TDPO are based on the implementation of DDPO-100. Additionally, due to the relatively small parameter size of our temporal critic, we opt for direct training on its entire parameter set. To isolate the effects of our neuron reset strategy, our TDPO-R mirrors TDPO in terms of concurrent components and configurations. All experiments were conducted on a system equipped with 8 NVIDIA A100 GPUs with 40GB of memory each.

\textbf{Hyperparameter configurations.} In Table~\ref{table:hyper}, we list the hyperparameter configurations for all implementations.

\begin{table*}[ht]
\caption{List of hyperparameter configurations for DDPO-2, DDPO-100, TDPO, and TDPO-R.}
\label{table:hyper}
\vskip 0.15in
\begin{center}
\begin{tabular}{l|c|c|c|c}
\toprule
Hyperparameters              & DDPO-2 & DDPO-100 & TDPO & TDPO-R \\
\midrule
Random seed                  & 42 & 42 & 42 & 42 \\
Denoising timesteps ($T$)    & 50 & 50 & 50 & 50 \\
Guidance scale               & 5.0 & 5.0 & 5.0 & 5.0 \\
Policy learning rate         & 3e-4 & 1e-4 & 1e-4 & 1e-4 \\
Policy clipping range        & 1e-4 & 1e-4 & 1e-4 & 1e-4 \\
Maximum gradient norm        & 1.0 & 1.0 & 1.0 & 1.0 \\
Optimizer                    & AdamW & AdamW & AdamW & AdamW \\
Optimizer weight decay       & 1e-4 & 1e-4 & 1e-4 & 1e-4 \\
Optimizer $\beta_1$          & 0.9 & 0.9 & 0.9 & 0.9 \\
Optimizer $\beta_2$          & 0.999 & 0.999 & 0.999 & 0.999 \\
Optimizer $\epsilon$         & 1e-8 & 1e-8 & 1e-8 & 1e-8 \\
Samples per epoch            & 256$\times T$ & 256$\times T$ & 256$\times T$ & 256$\times T$ \\
Training batch size          & 8 & 8 & 8 & 8 \\
Training steps per epoch     & 32$\times T$ & 32$\times T$ & 32$\times T$ & 32$\times T$ \\
Gradient accumulation steps  & 16$\times T$ & 16 & 16 & 16 \\
Gradient updates per epoch   & 2 & 2$\times T$ (100) & 2$\times T$ (100) & 2$\times T$ (100) \\
\midrule
Critic learning rate         & - & - & 1e-4 & 1e-4 \\
Critic clipping range        & - & - & 0.2 & 0.2 \\
\midrule
Neuron dormant threshold                   & - & - & - & 0 \\
Neuron reset frequency ($F$ / epochs)      & - & - & - & 10 \\
\bottomrule
\end{tabular}
\end{center}
\vskip -0.1in
\end{table*}




\section{Extended Analyses of Neuron States}
\label{appendix:neuron}

The following is an extension of the analyses regarding neuron states discussed in Section~\ref{section:4.2}. 

\begin{figure*}[ht]
\vskip 0.1in
\begin{center}
\centerline{
\includegraphics[width=0.45\textwidth]{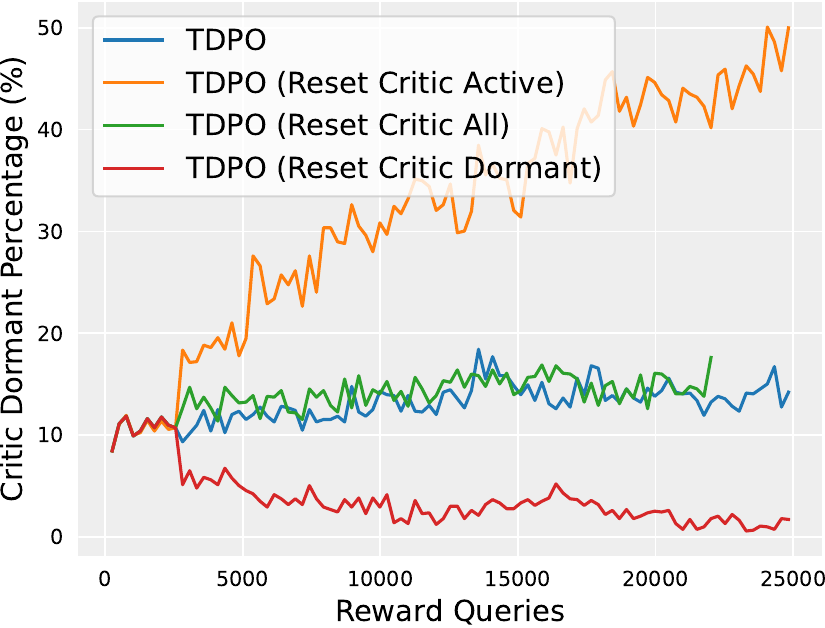}
\hskip 0.2in
\includegraphics[width=0.46\textwidth]{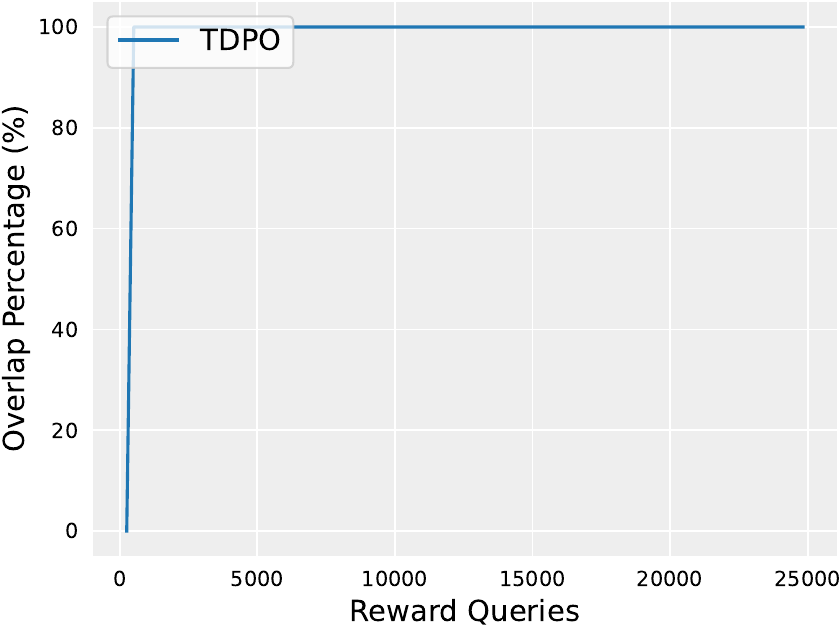}
}
\caption{The dormant percentage of neurons (left) and the overlap percentage between dormant neurons of current and previous training iterations (right) in our critic model when finetuning the diffusion model on Aesthetic Score.}
\label{dorm_c}
\end{center}
\vskip -0.2in
\end{figure*}

\textbf{Critic dormant neuron percentage.} In Section~\ref{section:4.2}, we described the effects of resetting different neurons in our critic model on the percentage of dormant neurons. Here we present the experimental results regarding these effects. The left plot in Figure~\ref{dorm_c} shows the percentage of dormant neurons in our critic model when finetuning the diffusion model on Aesthetic Score. As discussed in Section~\ref{section:4.2}, there is a slow ascent of the dormant percentage of neurons in our critic model during training. Resetting dormant neurons consistently reduces the dormant percentage, while resetting active neurons increases it significantly. This further substantiates a conclusion that resetting dormant neurons discourages the presence of dormant neurons, while resetting active neurons discourages the presence of active neurons.

\textbf{Overlap of dormant neurons.} 
To validate the persistence of dormant neurons throughout training, we track the overlap percentage between dormant neurons identified in current and previous training iterations.
The right plot in Figure~\ref{dorm_c} shows that the overlap percentage of dormant neurons in our critic model remains a value of 100 through the finetuning process. This indicates that, once a regularization with respect to dormant neurons is learned, then it will continuously affect subsequent training process. Combining this phenomenon with the empirical result we discussed in Section~\ref{section:4.2}, which is that resetting dormant neurons in our critic model exacerbates overoptimization, we can further extrapolate that dormant neurons in the critic model act as a adaptive regularization mechanism against overoptimization to imperfect rewards. While resetting dormant neurons may damage this implicit regularization, periodically resetting active neurons offers a potential mitigation strategy, encouraging the model to learn new regularization patterns without forgetting crucial past regularization.

\begin{figure*}[ht]
\vskip 0.2in
\begin{center}
\centerline{
\includegraphics[width=0.46\textwidth]{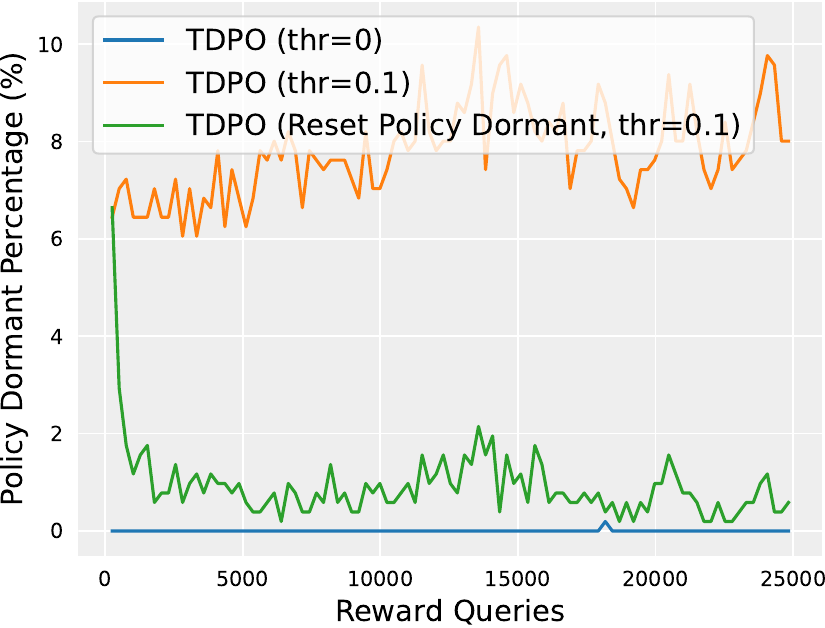}
\hskip 0.2in
\includegraphics[width=0.46\textwidth]{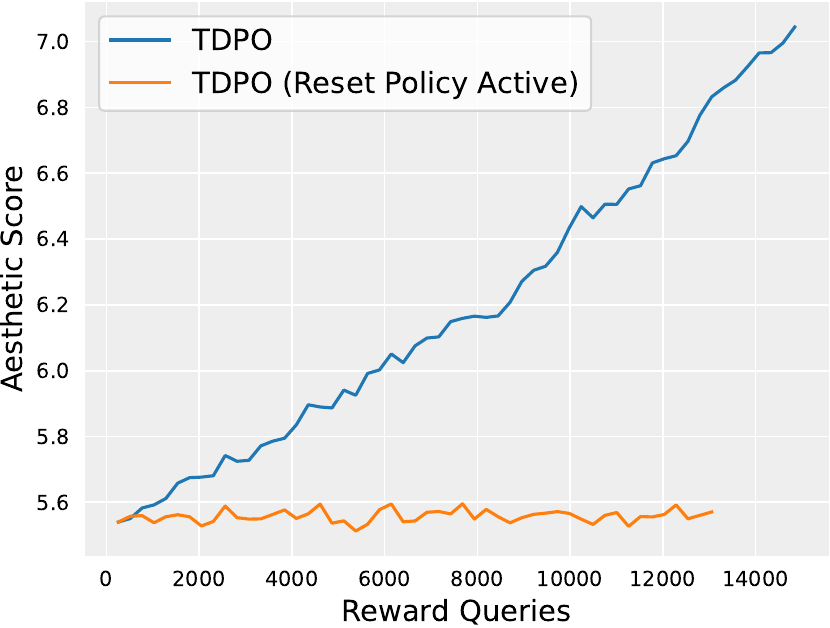}
}
\caption{The dormant percentage of neurons in the LoRA layers of our policy (left) and the effect of periodic resets of policy active neurons on the sample efficiency of TDPO (right) when finetuning the diffusion model on Aesthetic Score.}
\label{dorm_p}
\end{center}
\vskip -0.2in
\end{figure*}

\textbf{Policy dormant neuron percentage.} In Section~\ref{section:4.2}, we outlined the observation that a minimal number of dormant neurons is identified within the LoRA layers of our policy model.  Accordingly, the left plot in Figure~\ref{dorm_p} shows the dormant percentage of neurons in the LoRA layers of our policy when finetuned on Aesthetic Score. Notably, the plot indicates that the dormant percentage remains in close proximity to zero throughout the entire training duration when employing a dormant threshold of 0. Consequently, to discern a more pronounced proportion of dormant neurons, we elevated the threshold to 0.1, leading to higher dormant percentages during training. Despite the reduction in dormant neurons following the resets with the threshold of 0.1, discernible effects on cross-reward generalization are not prominently evident, as depicted in Figure~\ref{aes_img_reset}.

\textbf{Policy neuron reset.} In Section~\ref{section:4.2}, we described an effect that resetting active neurons in our policy model causes catastrophic forgetting and heavily hinders learning. Here we present the experimental result regarding this effect. The right plot in Figure~\ref{dorm_p} shows the effect of periodic resets of policy active neurons on the sample efficiency of TDPO when finetuning on Aesthetic Score. Compared to the original TDPO, the TDPO variant incorporating periodic resets of policy active neurons encounters substantial difficulty in optimizing the reward function, due to the fact that the resets of the overwhelming majority of the model parameters result in the loss of pre-learned knowledge. This unsatisfactory effect can be mitigated by replacing the online-updating scheme with an offline-updating scheme that incorporates a replay buffer to preserve prior knowledge and experiences. We highlight this replacement as an extension of our work for future research.

\begin{figure*}[ht]
\vskip 0.1in
\begin{center}
\centerline{
\includegraphics[width=0.48\textwidth]{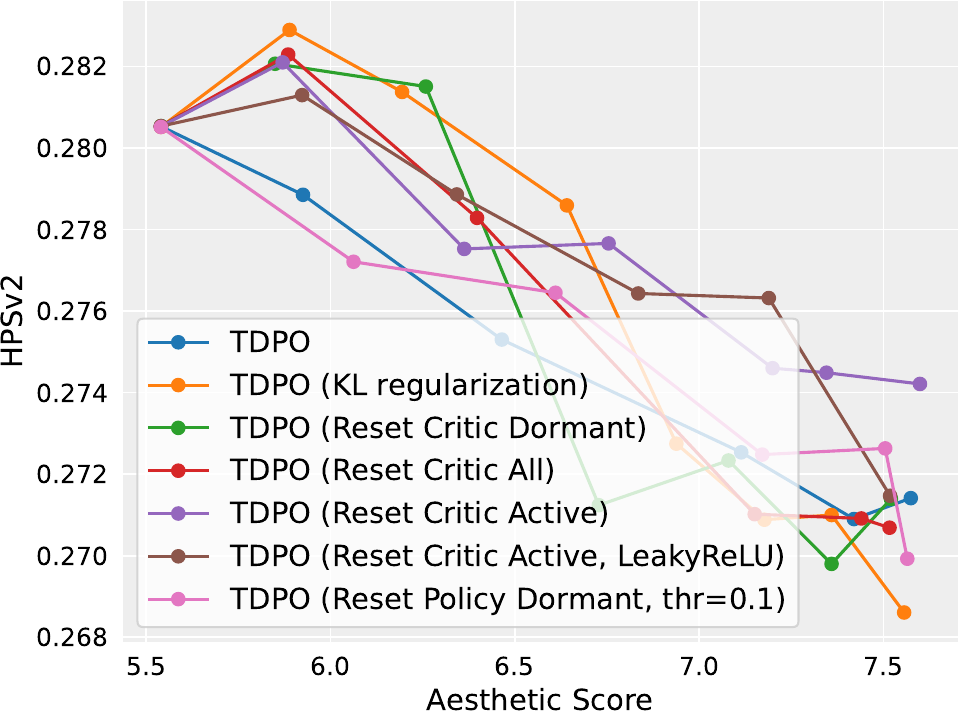}
\hskip 0.2in
\includegraphics[width=0.48\textwidth]{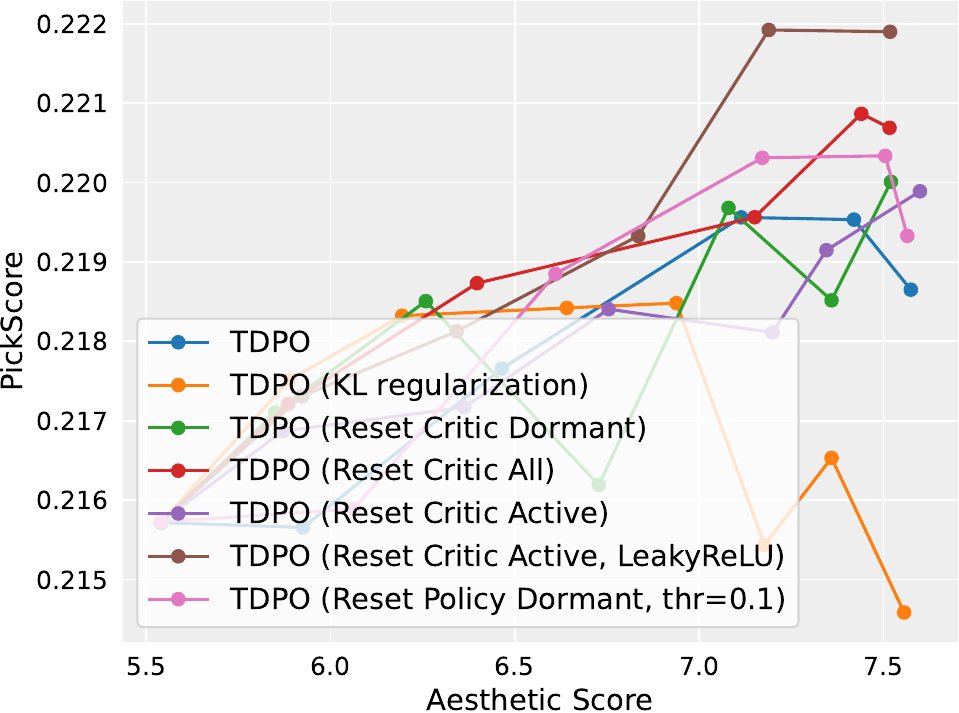}
}
\caption{Additional results of cross-reward generalization against HPSv2 (left) and PickScore (right) when finetuning the diffusion model on Aesthetic Score, comparing different variants of TDPO.}
\label{reset_app}
\end{center}
\vskip -0.2in
\end{figure*}

\textbf{Additional results of cross-reward generalization.} Here is an extension of the cross-reward generalization results for different variants of TDPO presented by Figure~\ref{aes_img_reset} in Section~\ref{section:5.3}. In Figure~\ref{reset_app}, we show more cross-reward generalization results for TDPO variants with different neuron reset strategies or the KL regularization mechanism. The diffusion models in all variants are finetuned on Aesthetic Score and evaluated on HPSv2 and PickScore. We further investigate the cross-reward generalization capability of TDPO-R by employing an alternative evaluation with leaky ReLU \cite{leakyrelu} instead of the standard ReLU, achieving an even superior performance when evaluating cross-reward generalization against PickScore.

\section{Generalization to Unseen Prompts}
\label{appendix:unseen}

In order to further validate the effectiveness and robustness of our methods, here we extend the cross-reward evaluation to new text prompts that are not previously seen by models during the finetuning process.

\begin{figure*}[ht]
\vskip 0.1in
\begin{center}
\centerline{
\includegraphics[width=0.325\textwidth]{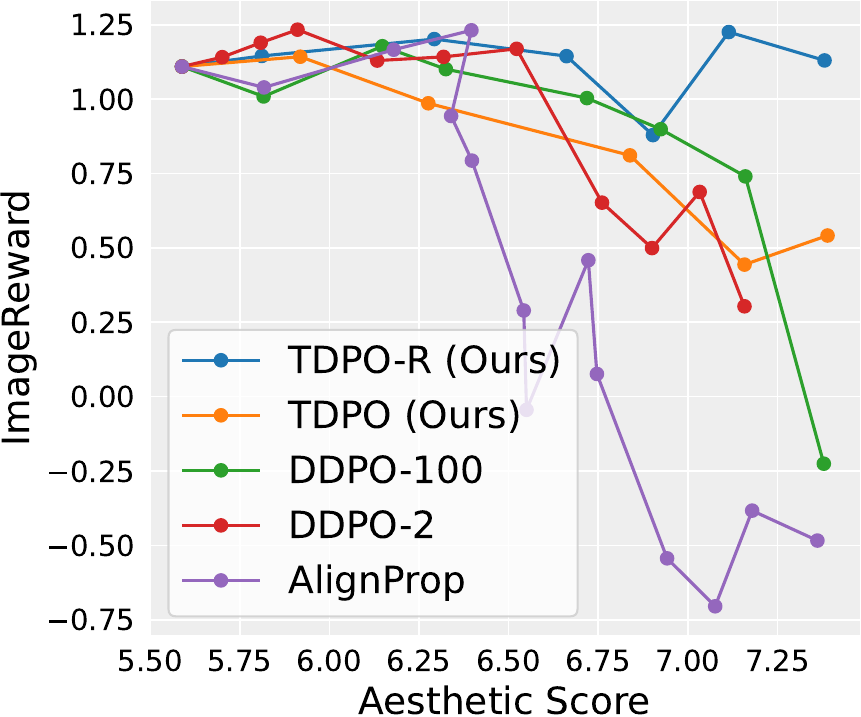}
\hskip 0.05in
\includegraphics[width=0.325\textwidth]{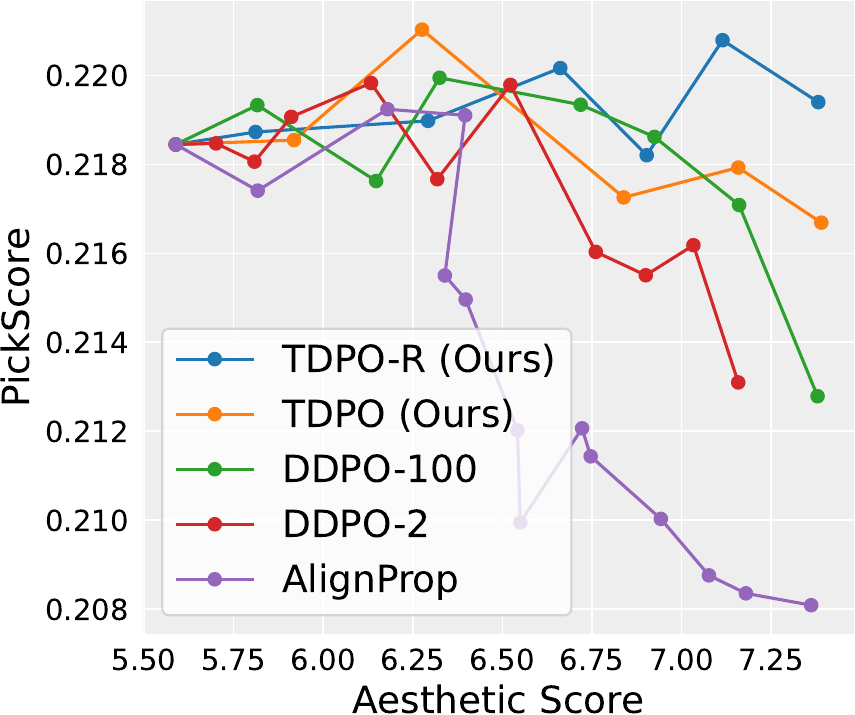}
\hskip 0.05in
\includegraphics[width=0.325\textwidth]{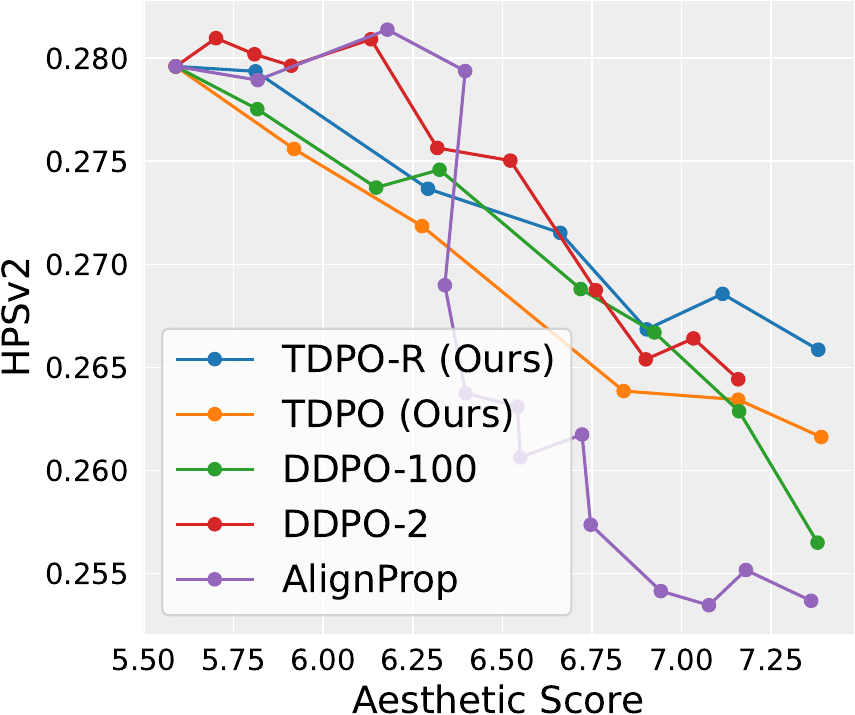}
}
\caption{Cross-reward generalization results evaluated on a text prompt set of unseen animals when finetuning the diffusion model on Aesthetic Score.}
\label{unseen_appendix}
\end{center}
\vskip -0.2in
\end{figure*}

\textbf{Unseen animals.} We first employ a novel text prompt set consisting of 8 unseen animal names, including ``snail", ``hippopotamus", ``cheetah", ``crocodile", ``lobster", ``octopus", ``elephant", and ``jellyfish". We conduct evaluations of cross-reward generalization over samples generated using these unseen animal prompts during the finetuning process. In Figure~\ref{unseen_appendix}, we show the evaluation results when finetuning the diffusion model on Aesthetic Score via AlignProp, DDPO-2, DDPO-100, as well as our TDPO and TDPO-R. Notably, our TDPO and TDPO-R still maintain superior performance in cross-reward generalization compared to DDPO and AlignProp. These out-of-domain evaluations demonstrate the robust capabilities of TDPO-R in mitigating reward overoptimization, generalizing effectively to out-of-domain prompts.

\begin{figure*}[ht]
\vskip 0.1in
\begin{center}
\centerline{
\includegraphics[width=0.32\textwidth]{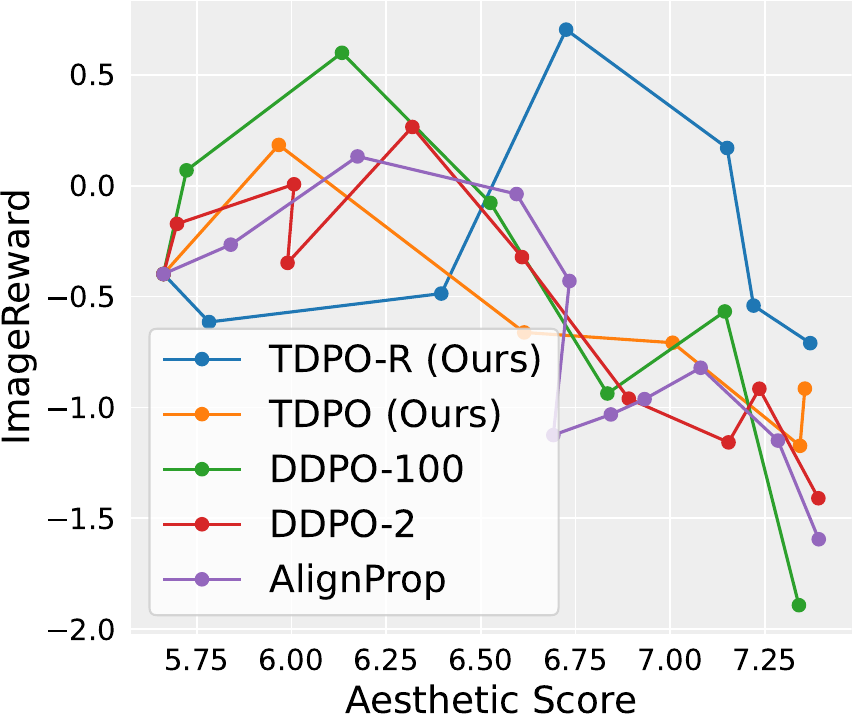}
\hskip 0.05in
\includegraphics[width=0.33\textwidth]{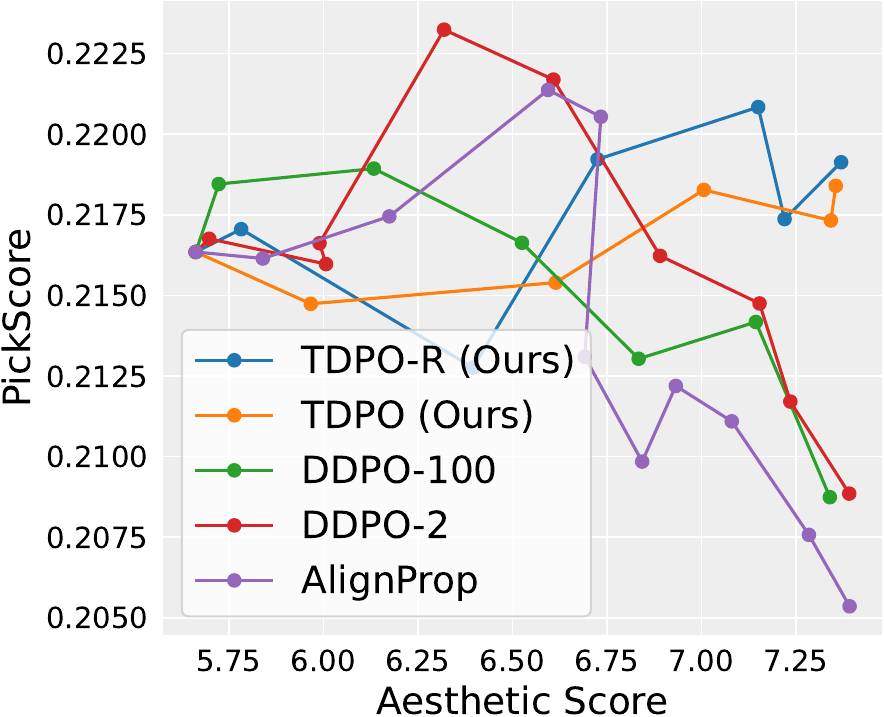}
\hskip 0.05in
\includegraphics[width=0.325\textwidth]{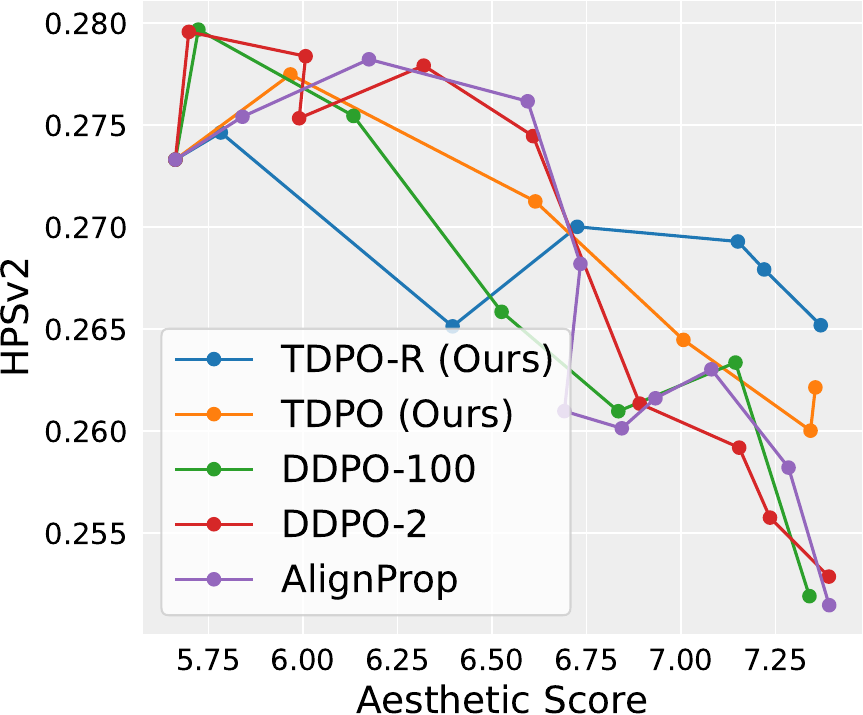}
}
\caption{Cross-reward generalization results evaluated over unseen text prompts involving color (``A green colored rabbit"), count (``Four wolves in the park"), composition (``A cat and a dog"), and location (``A dog on the moon") while finetuning on Aesthetic Score.}
\label{complex}
\end{center}
\vskip -0.2in
\end{figure*}

\textbf{Color, count, composition, and location.} Furthermore, we adopt a set of complex text prompts involving specific color (``A green colored rabbit"), count (``Four wolves in the park"), composition (``A cat and a dog"), and location (``A dog on the moon") as introduced in \cite{dpok}. In \cite{dpok}, these prompts are originally used as training text prompts, meaning that they are seen by models during the finetuning process. In contrast, we utilize them as unseen prompts for cross-reward evaluations while finetuning models on Aesthetic Score, as illustrated in Figure~\ref{complex}. 

\section{Additional Qualitative Results}
\label{appendix:vis}

\begin{figure*}[ht]
\vskip 0.2in
\begin{center}
\centerline{
\includegraphics[width=\textwidth]{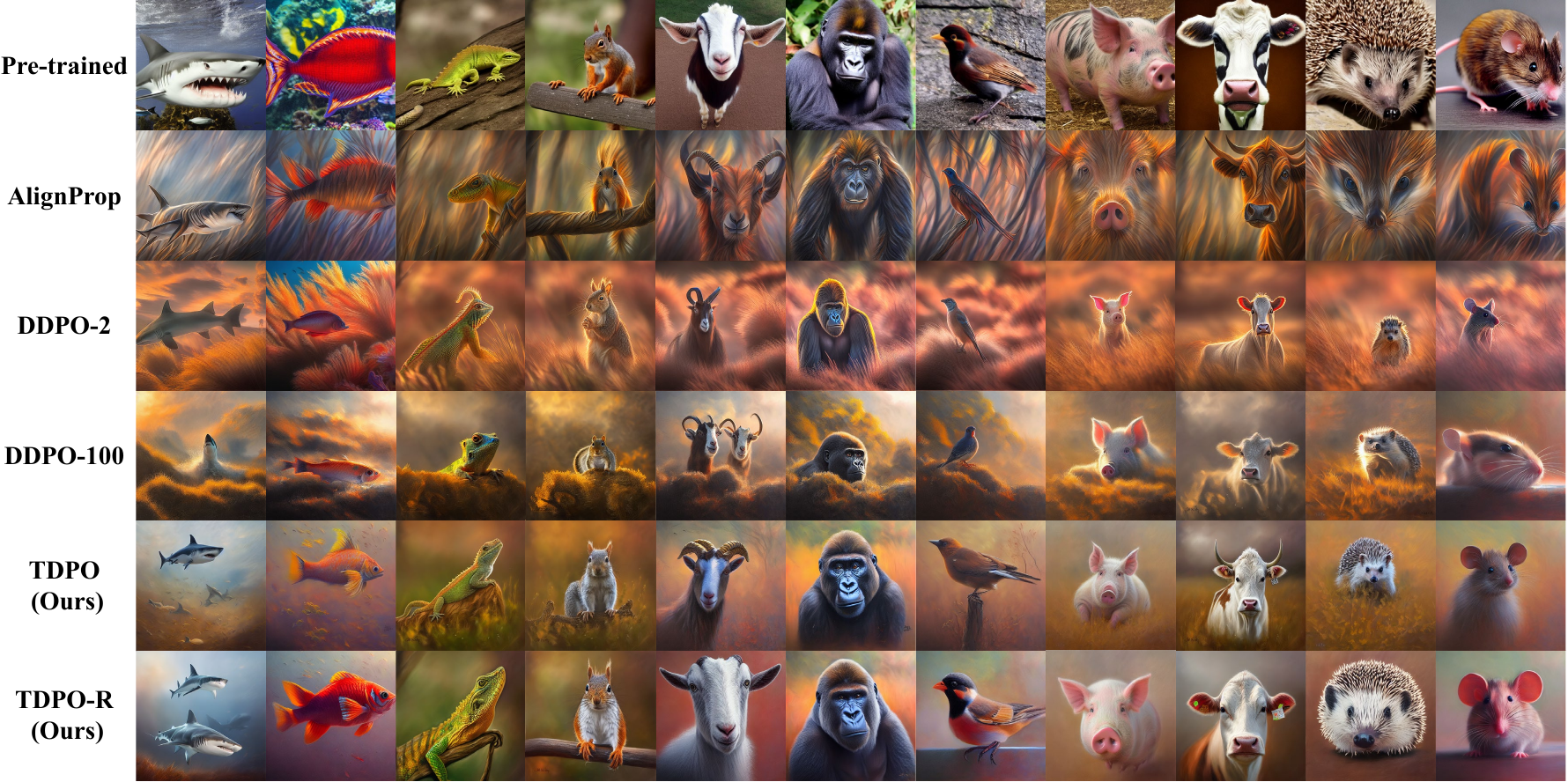}
}
\caption{Additional qualitative results sampled from models that are either pre-trained or further finetuned on Aesthetic Score via AlignProp, DDPO-2, DDPO-100, as well as our TDPO and TDPO-R. For a fair comparison, all images are generated using a fixed random seed of 42. Additionally, for the fine-tuned models, the aesthetic scores of the generated images achieve similar values around 7 $\pm$ 0.1.}
\label{vis_appendix}
\end{center}
\vskip -0.2in
\end{figure*}

Here is an extension of the qualitative results presented by Figure~\ref{vis} in Section~\ref{section:5.4}. In Figure~\ref{vis_appendix}, we present additional qualitative results with high-reward images on Aesthetic Score. The results from AlignProp and DDPO show notable saturation in terms of style, background, and sunlight, while our generation results manifest greater diversity in these aspects. Specifically, AlignProp generates images characterized by a fixed painting style, while the results from DDPO exhibit a photographic style with similar sunlight angles and similar grassy backgrounds, even in response to prompts like ``shark" and ``fish". Conversely, our TDPO and TDPO-R demonstrate the capacity to generate images encompassing both painting and photographic styles, and exhibit an enhanced proficiency in generating diverse and coherent backgrounds aligned with given prompts. In our interpretation, Aesthetic Score characterizes a preference for images that exhibit a stylistic amalgamation, comprising elements reminiscent of both painting and photography. Accordingly, our algorithms ensure effective optimization towards this preference against overfitting a fixed style.

\section{Extended Analysis of Encoder Alignment in Temporal Critic}
\label{appendix:encoder}

\textbf{Reward model encoders.} Following established practices \cite{aes,ddpo}, we leverage a pre-trained CLIP model \cite{clip} as the encoder of the reward model for Aesthetic Score. For the HPSv2 and PickScore reward models, we adopt their official PyTorch implementations, each of which utilizes a customized OpenCLIP-H model \cite{openclip} finetuned on their specific preference data as the encoder. This ensures consistency with the established rewarding procedures associated with these models.

\textbf{Temporal critic encoders.} As introduced in \ref{section:4.1}, while finetuning on a specific reward function (Aesthetic Score, HPSv2, or PickScore), we incorporate the corresponding encoder from the respective reward model into our temporal critic. This encoder extracts embeddings from the decoded images w.r.t. each intermediate latent feature across all timesteps of the denoising process. These embeddings serve as the input to a lightweight Multi-Layer Perceptron (MLP) containing only 5 linear layers, with progressively decreasing output dimensionalities of 1024, 128, 64, 16, and 1 unit in the final layer. Crucially, this practice of reusing encoders establishes an alignment between the encoders of both the reward model and the temporal critic. This alignment tends to be critical for overall performance, as it ensures consistency in feature representations and enables the temporal critic to inherit the inductive bias of the reward model during initial training. Beyond performance gains, encoder alignment also offers a compelling advantage in memory efficiency, as the need to store a separate encoder for the temporal critic is eliminated, especially for large pre-trained models like CLIP \cite{clip} and OpenCLIP-H\cite{openclip}.

\begin{figure*}[ht]
\vskip 0.2in
\begin{center}
\centerline{
\includegraphics[width=0.33\textwidth]{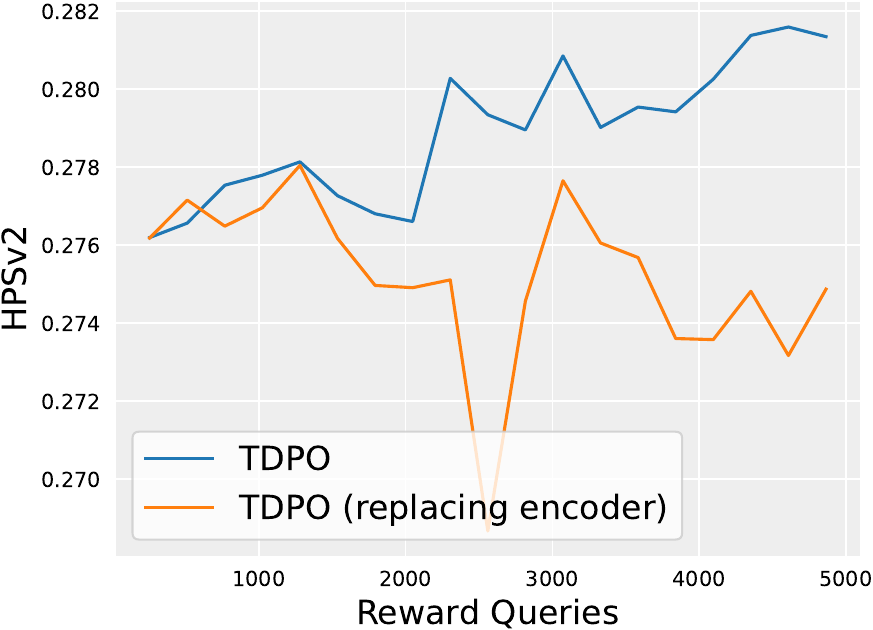}
\includegraphics[width=0.33\textwidth]{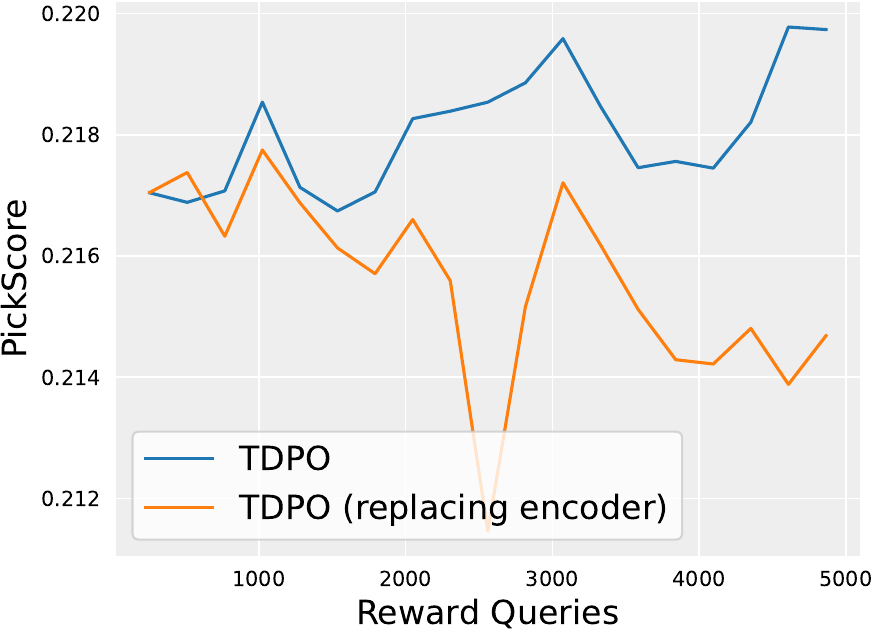}
\includegraphics[width=0.33\textwidth]{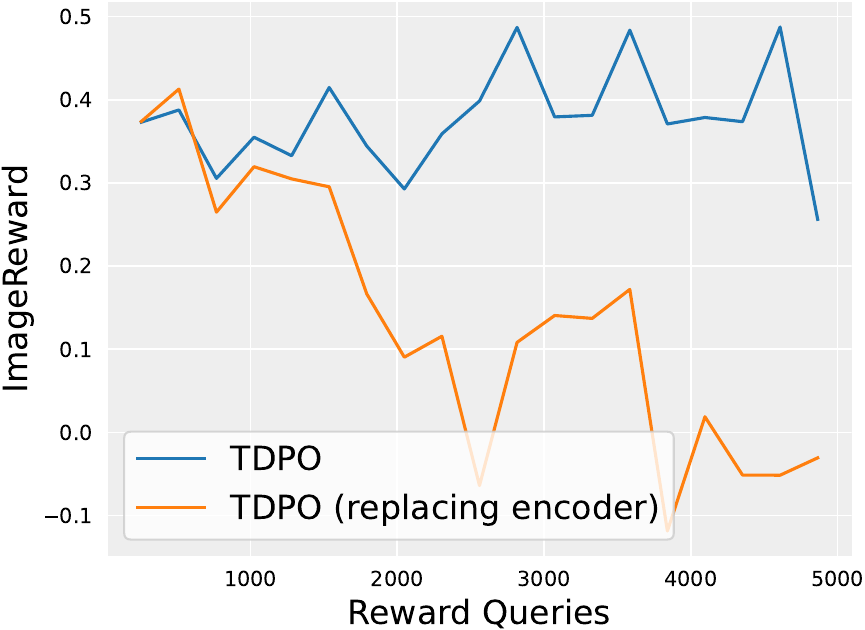}
}
\caption{Evaluation results of TDPO's reward optimization performance when finetuning the diffusion model on HPSv2 and replacing the HPSv2 encoder in the temporal critic with a misaligned encoder from Aesthetic Score.}
\label{encoder}
\end{center}
\vskip -0.2in
\end{figure*}

\textbf{Impact of misaligned encoders.} To delve deeper into the impact of misaligned encoders, we conduct an additional experiment where we replace the HPSv2 encoder in the temporal critic with a misaligned encoder from Aesthetic Score while finetuning the diffusion model on HPSv2. As illustrated in Figure~\ref{encoder}, this misalignment of the encoders lead to a significant decline in TDPO's reward optimization performance. This finding highlights the critical role of encoder alignment between the reward model and the temporal critic for effective reward finetuning, as discrepancies in feature representations can hinder the critic's ability to guide optimization towards the desired reward.

\section{Extended Related Work}
\label{appendix:work}

\textbf{Generation control of diffusion models.} Following prior works \cite{ddpo, draft} on diffusion model alignment, we incorporate Classifier-Free Guidance (CFG) \cite{cfg} to perform conditional generation of diffusion models. Prior works \cite{ddpo, draft} provided compelling evidence that the alignment methods with CFG-based generation control outperform other approaches including prompt engineering and classifier guidance \cite{Dhariwal2021,Bansal2023}. Accordingly, to maintain clarity and emphasize our improvements in diffusion model alignment, we refrain from conducting comparative analyses on various generation control techniques.

\textbf{Reinforcement learning for diffusion models.} \citet{sftpg} first utilize reinforcement learning to train diffusion models. Subsequent studies by \citet{dpok} and \citet{ddpo} delve into the utilization of policy gradient-based algorithms to align text-to-image diffusion models with arbitrary reward functions. Instead of finetuning model parameters, \citet{hao2023} apply reinforcement learning to optimize prompts for text-to-image diffusion models. \citet{wang2023} leverage diffusion models to create policies for offline reinforcement learning beyond the context of text-to-image generation.


\end{document}